\documentclass[manuscript]{acmart}

\AtBeginDocument{%
  }

\setcopyright{acmlicensed}
\copyrightyear{2026}
\acmYear{2026}
\acmDOI{XXXXXXX.XXXXXXX}
\acmISBN{978-1-4503-XXXX-X/2018/06}

\title[Experts Rise Where LLMs Disagree]{Experts Rise Where LLMs Disagree: Using Cross-Model Disagreement to Target Expert Effort in LLM Codebook Revision for Large-Scale Annotation}

\usepackage{xspace}
\usepackage{comment}
\usepackage{multirow}
\usepackage{subcaption}
\usepackage{threeparttable}

\usepackage{enumitem}
\usepackage{amsmath}
\usepackage{booktabs}
\usepackage[dvipsnames]{xcolor}
\usepackage{hyperref}

\usepackage[most]{tcolorbox}
\usepackage{listings}
\usepackage{needspace}   

\lstdefinestyle{prompt}{
  basicstyle=\ttfamily\scriptsize,
  breaklines=true, breakatwhitespace=false,
  columns=fullflexible, keepspaces=true,
  showstringspaces=false, upquote=true,
}

\newtcblisting[auto counter]{promptlisting}[1]{
  listing only, breakable, enhanced,
  colback=gray!3, colframe=gray!55,
  boxrule=0.4pt, arc=2pt, left=3pt, right=3pt, top=2pt, bottom=2pt,
  fonttitle=\small\bfseries, title={#1},
  listing options={breaklines=true, style=prompt},
}

\begin{document}


\author{Zeyu He}
\affiliation{%
  \institution{The Pennsylvania State University}
  \city{University Park}
  \state{PA}
  \country{USA}
}
\email{zmh5268@psu.edu}

\author{Zhuqian Zhou}
\affiliation{
  \institution{Cornell University}
  \city{Ithaca}
  \state{NY}
  \country{USA}
}
\email{zz968@cornell.edu}

\author{Kirk Vanacore}
\affiliation{
  \institution{Cornell University}
  \city{Ithaca}
  \state{NY}
  \country{USA}
}
\email{kpv27@cornell.edu}

\author{Rene F. Kizilcec}
\affiliation{
  \institution{Cornell University}
  \city{Ithaca}
  \state{NY}
  \country{USA}
}
\email{kizilcec@cornell.edu}

\author{Ting-Hao `Kenneth' Huang}
\affiliation{%
  \institution{The Pennsylvania State University}
  \city{University Park}
  \state{PA}
  \country{USA}
}
\email{txh710@psu.edu}








\renewcommand{\shortauthors}{He et al.}




\begin{abstract}
Large-scale text annotation brings expert insight to millions of documents, often through a codebook that AI annotators follow. Developing a robust codebook, however, takes months. Large language models (LLMs) could speed this process by applying an early codebook to the data, surfacing cases with strong LLM disagreement, and eliciting expert feedback to address them. We examined three ways experts can provide feedback for LLM codebook revision: {\em (i)} editing LLM-generated revisions driven by cross-LLM disagreement (Codebook Verifying), {\em (ii)} answering questions about LLM disagreements (Question Answering), and {\em (iii)} labeling disagreement cases with rationales (Rationale Labeling). Experiments on thousands of tutoring-session transcripts show that \textbf{Rationale Labeling yielded the highest LLM-labeling accuracy (64.9\%) against expert labels, outperforming the expert-revised codebook (57.8\%)}. The best Question Answering setting also outperformed it (60.5\%). Our work shows that LLMs can be used to strategically target expert attention, shortening months of codebook revision to days without sacrificing labeling performance.
\end{abstract}

\maketitle




\section{Introduction}

Large-scale data annotation brings expert insights to thousands or millions of documents.
For example,
linguists encoded grammatical expertise into annotation schemes, which annotators then scaled across nearly 200 languages~\cite{nivre-etal-2016-universal};
to study religious themes in fiction, literature scholars developed a codebook for identifying ``acts of God'', then used large language models (LLMs) to scale it to over 20,000 passages across 88 novels~\cite{10.63744@U7SMpKE7aJs1};
to make research methods and contributions in COVID-19 papers easier to find, 
researchers used crowdsourcing to apply an expert-developed codebook to over 100,000 text segments from 10,966 papers~\cite{huang-etal-2020-coda,10.1145/3613904.3642834};
political scientists used LLMs to apply expert-developed codebooks 
to over 48,000 political texts, identifying events such as protests and political violence~\cite{halterman2026codebook}; and, 
at a much larger scale, 
the \textsc{CAPC-CG} project used LLMs to scale a codebook for policy directives to 3.3 million Chinese government policy paragraphs~\cite{sun-etal-2026-capc}.
Across these projects, expert insights were externalized and crystallized into written \textbf{codebooks} that human or AI annotators can follow.

However, developing a robust codebook that defines the labels and rules for applying them requires substantial expert effort.
To name just a few examples from the projects above, in the literature study focusing on religious themes, scholars spent two semesters developing the codebook, meeting every week to examine difficult cases, discuss disagreements, and revise the codes~\cite{10.63744@U7SMpKE7aJs1}.
In the COVID-19 paper project, a biomedical expert spent substantial time manually annotating data to identify corner cases and shape the coding instructions~\cite{huang-etal-2020-coda,10.1145/3613904.3642834}. 
Even in the LLM-based \textsc{CAPC-CG} project, three experts spent three months refining the codebook: each week, they independently labeled about 500 policy paragraphs, compared disagreements, and revised the coding rules~\cite{sun-etal-2026-capc}.

LLMs' ability to take in large amounts of text and act accordingly opens possibilities for speeding up this process.
Our intuition is that \textbf{strong disagreement across multiple LLMs on how to label an instance may signal a problem in the codebook, such as an ambiguity, a missing rule, or a difficult boundary}. 
This intuition builds on evidence that LLMs outperform average online workers on many text annotation tasks~\cite{10.1145/3613904.3642834},
suggesting that their labels carry meaningful signals even when they remain less reliable than expert labels.

In this paper, we examine this idea in a large-scale annotation project that aims to identify and analyze tutoring strategies across thousands of tutoring-session transcripts.
This project was structured as a traditional expert-led data annotation project: 
two experts
first quickly put together an initial codebook based on their knowledge and an initial review of the data. 
They then met weekly on Zoom to read and annotate data, discuss disagreements, and revise the codebook over a period of six months.
In our experiments, we took the earliest version of the codebook and had six different LLMs use it to annotate 6,595 utterances that the experts had never seen. 
From these LLM-annotated data, we identified cases where the LLMs strongly disagreed, then had an LLM use these disagreement cases to revise the codebook through three different ways for experts to provide input: 
{\em (i)} \textbf{Codebook Verifying (CV)}, in which experts edited LLM-generated revisions based on disagreement cases; 
{\em (ii)} \textbf{Question Answering (QA)}, in which experts answered LLM-generated questions about disagreement cases, and an LLM used their answers to revise the codebook; and 
{\em (iii)} \textbf{Rationale Labeling (RL)}, in which experts labeled disagreement cases and explained their reasoning, and an LLM used these labels and rationales to revise the codebook.
We evaluated the resulting codebooks from each condition, as well as the expert-revised codebook that took six months to develop, against the expert annotations produced during that period. 
We found that
\textbf{Rationale Labeling (RL) produced the highest LLM-labeling accuracy (64.9\%) and weighted F1 score (0.658), outperforming the expert-revised codebook (accuracy = 57.8\%; F1 = 0.583).} 
Question Answering (QA), with its best setting, also outperformed the expert-revised codebook by a smaller margin (accuracy = 60.5\%; F1 = 0.613). 
These results show that targeted feedback from experts familiar with the task can substantially improve an initial codebook's effectiveness for LLM annotation with a few additional hours of expert effort.

Our contribution is a workflow that turns model disagreement into targeted expert input and then into reusable codebook instructions. We contribute three interfaces for eliciting that input, an empirical comparison of the resulting codebooks in a longitudinal tutoring-annotation project, and evidence about their annotation performance, recorded expert effort, and perceived demands. The results motivate interfaces that help experts express difficult judgments while allowing LLMs to translate those judgments into explicit annotation rules.

\begin{figure*}[t]
    \centering
    \includegraphics[width=0.99\linewidth]{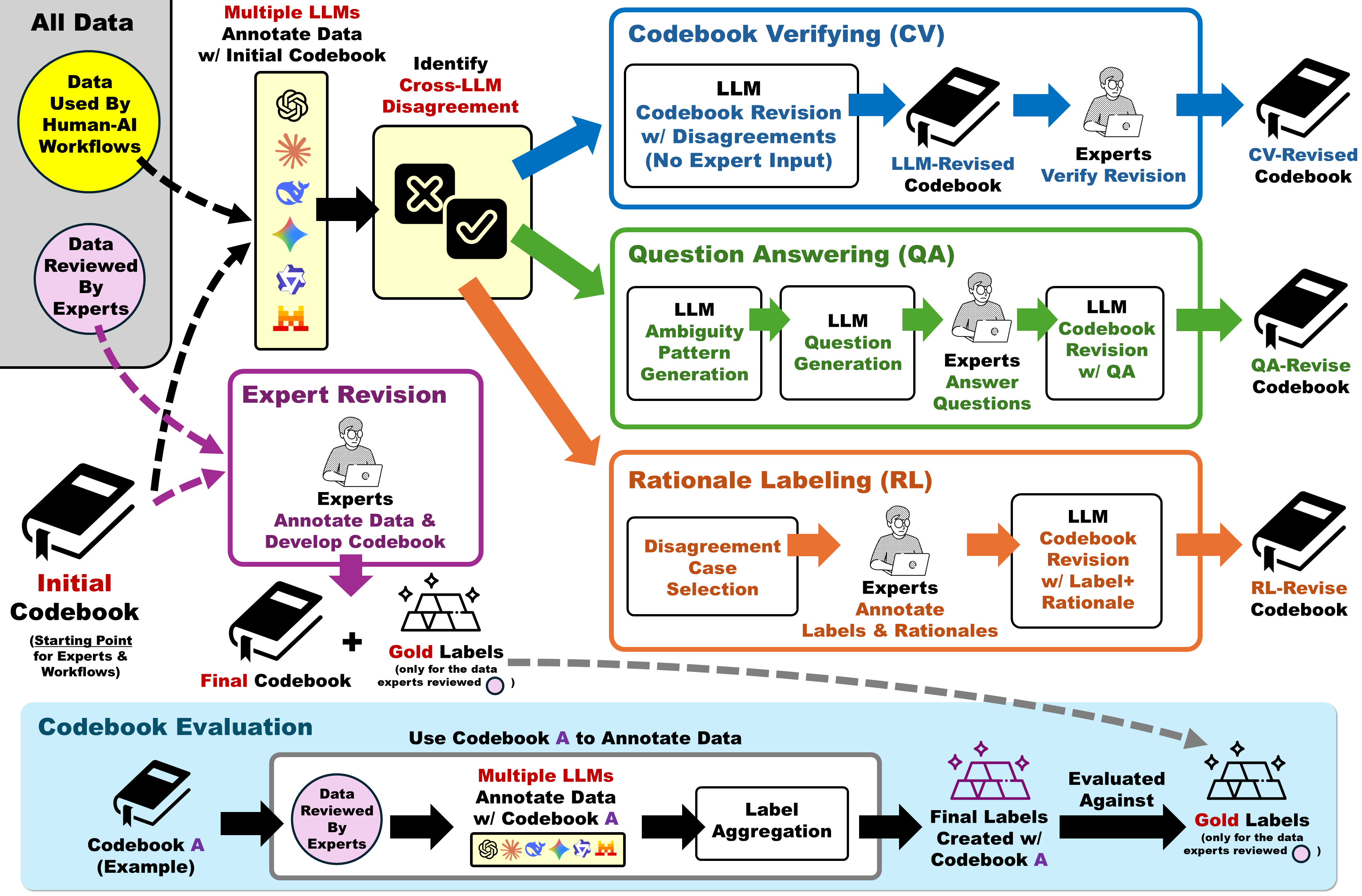}
    \caption{Overview of the codebook revision and evaluation workflow. All approaches begin with the same Initial Codebook but use separate data for expert-driven codebook development and human-AI revision. In the expert-driven process, experts annotate data, resolve disagreements, and iteratively revise the codebook, producing the Final Codebook and expert-consensus gold labels. In the three human-AI workflows, multiple LLMs annotate a shared revision corpus using the Initial Codebook, and cross-LLM disagreement guides expert effort. In Codebook Verifying (CV), an LLM proposes codebook revisions from disagreement evidence, which experts review and edit. In Question Answering (QA), an LLM identifies ambiguity patterns, generates clarification questions, and uses expert answers to revise the codebook. In Rationale Labeling (RL), experts label selected disagreement cases and provide rationales that guide LLM revision. The bottom panel illustrates evaluation using Codebook A as an example: multiple LLMs annotate the same expert-reviewed test set with each codebook; their predictions are aggregated into final labels, and these labels are compared with expert consensus gold labels to compute accuracy and weighted F1.}
    \label{fig:workflow}
    \Description{Flowchart showing codebook revision and evaluation. All workflows begin with the same initial codebook. Using one dataset, experts annotate data, discuss disagreements, and iteratively produce the final codebook and expert-consensus gold labels. Using a separate dataset, multiple LLMs identify annotation disagreements that inform three human–AI workflows: Codebook Verifying (CV), where experts review and edit LLM-proposed revisions; Question Answering (QA), where experts answer clarification questions generated from ambiguity patterns; and Rationale Labeling (RL), where experts label disagreement cases and provide rationales. Expert input guides the revised codebook produced by each workflow. During evaluation, multiple LLMs annotate a common evaluation set using each codebook. Their predictions are aggregated and compared with expert-consensus gold labels to measure accuracy and weighted F1.}
\end{figure*}

\subsection{Research Questions\label{sec:research-questions}}
In this paper, we aim to answer three research questions in the context of large-scale data annotation:

\begin{enumerate}[label=\textbf{RQ\arabic*:}]
    
\item
How do codebooks produced by the three human-AI revision workflows compare with the initial and final expert-developed codebooks in LLM annotation performance?

\item
What tradeoffs do the workflows show between annotation performance, recorded expert effort, and experts' reported experiences?

\item
How is cross-model disagreement associated with annotation difficulty at the utterance and label levels?

\end{enumerate}

Answers to these questions can inform future expert-led large-scale data annotation projects.

\label{sec:introduction}

\section{Related Work}




\subsection{Codebooks for LLM Annotators}
As LLMs become the annotators in large-scale projects, the codebook becomes the primary lever for controlling annotation quality. Many projects have already replaced human annotators with LLMs in areas that were traditionally the province of people, crowdsourced labeling and qualitative coding among them, and in each case an expert-written codebook carried the expertise that the human annotators used to supply~\cite{10.1145/3613904.3642834,xiao2023codebook,halterman2026codebook}.
Halterman and Keith~\cite{halterman2026codebook} showed that LLM measurements of political science concepts, such as protest or political violence, depend heavily on how a concept was specified, and that supplying the actual codebook text---rather than a label name---materially changed validity. 
He et al.~\cite{10.1145/3613904.3642834} studied where GPT-4 could and could not substitute for crowd workers in an annotation pipeline built around an expert codebook, and Xiao et al.~\cite{xiao2023codebook} paired an expert codebook with an LLM for deductive coding. 
These projects treated the codebook as an input produced through conventional expert deliberation, which can require substantial time.



Recently, several projects attempted to use human-AI approaches to speed up or improve this codebook and prompt iteration process. 
He et al.~\cite{he2025promptingdark} studied ``prompting in the dark,'' in which users iteratively revised prompts to label data with no gold-standard benchmark to measure progress against; across 20 participants, the process proved highly unreliable, with only 9 improving labeling accuracy after four or more iterations. 
Zamfirescu-Pereira et al.~\cite{zamfirescu2023johnny} similarly found that people without AI expertise struggled to convert their intent into effective prompts, tending to over-generalize from single failures. 
These results did not yet offer a promising route to faster codebook development, and they left open the question this paper takes up: 
what kind of expert input, directed at which cases, actually improves a codebook when gold labels are scarce.

\subsection{Automatic Instruction or Prompt Optimization}
A parallel line of work treated the instruction itself as a parameter to be optimized automatically. 
Automatic Prompt Optimization (APO)~\cite{pryzant2023automatic} used textual ``gradients'' derived from failure cases; APE~\cite{zhou2023ape} and OPRO~\cite{yang2024opro} searched instruction space with an LLM proposer; DSPy~\cite{khattab2024dspy} and TextGrad~\cite{yuksekgonul2025textgrad} generalized this to compiled multi-stage pipelines. 
All of these depended on a reliable objective, which in annotation meant a reasonable quantity of gold labels. 
Self-Supervised Prompt Optimization (SPO)~\cite{xiang2025self} relaxed that requirement by substituting LLM self-evaluation, but still assumed the model can tell better outputs from worse ones. 
Real-world annotation projects, however, frequently satisfy neither assumption. Codebooks in these projects often carry dozens of fine-grained categories rather than a handful; 
ground truth is not a lookup but the product of two or more domain experts annotating independently and then negotiating a consensus; 
and gold labels accumulate slowly as the codebook evolves, 
so the early stages---exactly when revision matters most---have almost none. 
Where the categories turn on distinctions that experts themselves had to argue out, an LLM's self-assessment is unlikely to substitute for the missing objective. 
Our study results echo this empirically: under a no-gold-label constraint, both APO and SPO fell below the unrevised initial codebook. 

\subsection{Human-AI Approaches to Aligning AI Annotator Behavior}
\label{sec:related-work}

In response to the high cost of building a good codebook, another line of recent work tried to bypass the codebook altogether and align AI annotators' behavior directly. 
Wang et al.~\cite{wang2024verification} had experts verify LLM labels before those labels enter the dataset, 
and Shankar et al.~\cite{shankar2024validators} had experts iteratively refine evaluation criteria by grading LLM outputs.
Both aligned experts with a model's judgments on individual items.

We instead route limited expert judgment into a durable, human-readable codebook. 
Unlike per-item alignment, this allows expert decisions to guide many future cases without keeping experts continuously in the loop. 
Codebooks also preserve an inspectable representation of expert consensus that can transfer across data and models.


\subsection{Instruction Design for Human Annotators}
Also relevant is the crowdsourcing literature, which has long treated task instructions as a first-class design object.
Sprout~\cite{bragg2018sprout} elicited worker questions to expose gaps in task descriptions; WingIt~\cite{manam2018wingit} routed worker-reported ambiguities back to requesters for clarification; and work on task clarity showed that instruction quality predicted output quality~\cite{gadiraju2017clarity}. 
Closest in spirit to our approach, Revolt~\cite{chang2017revolt} used disagreement among crowd workers to surface ambiguous concepts and sharpen label definitions, and structured labeling~\cite{kulesza2014concept} supported the evolution of a concept as annotators encounter new cases.
Deliberative designs such as MicroTalk~\cite{drapeau2016microtalk} improved accuracy by having workers argue over contested items. 
These systems directly inspired our approach: like Revolt, we treat annotator disagreement as a signal about the label definition rather than as noise to be averaged away. 
In our work, instead of online crowd workers, we used LLM annotators to process large amounts of text data faster. 


\section{LLM-Based Codebook Revision Using Expert Feedback Elicited from Cross-Model Disagreement\label{sec:method}}
We propose a workflow in which multiple LLMs first annotate a large dataset using an early codebook (Figure~\ref{fig:workflow}).
The workflow then identifies instances with strong cross-model disagreement (Section~\ref{sec:disagreement-method}) and elicits expert input on these cases through 
Codebook Verifying (CV) (Section~\ref{sec:CV-method}),
Question Answering (QA) (Section~\ref{sec:QA-method}), or 
Rationale Labeling (RL) (Section~\ref{sec:RL-method}).
Finally, an LLM uses the expert input to revise the codebook. 

\subsection{Target Scenario and Design Constraints\label{sec:design-constraint}}
We target long-term data annotation projects that aim to scale expert insights beyond what experts can manually annotate. 
In these projects, a small group of domain experts reads and annotates data, discusses disagreements, and iteratively develops a codebook. 
As prior work shows, this process can take months. 
Once the codebook is developed, LLMs can apply it at a much larger scale.
We also assume that these experts have specialized domain knowledge and can annotate the data more reliably than state-of-the-art LLMs. 

Our target scenario introduces three constraints. 
First, \textbf{expert time is valuable and limited}. 
We therefore assume that experts can provide only a limited amount of input rather than spend countless hours reviewing data. 
Second, we \textbf{do not assume access to a large collection of gold labels}.
LLM-based annotation projects often begin with few or no gold labels and develop them gradually as the codebook evolves. 
Third, we assume \textbf{access to an early codebook} that LLMs can use to begin annotating the data.

\subsection{Preparation Phase: Multi-LLM Annotation and Cross-Model Disagreement Calculation\label{sec:disagreement-method}}
The foundation of our workflow is cross-model disagreement on unlabeled data. 
Before eliciting expert feedback or revising the codebook, we first ask multiple LLMs to independently annotate a collection of data instances using an initial codebook that experts developed early in the project. 
We refer to each data instance as an \textit{utterance}.

We use LLMs from different model providers to encourage diversity in their predictions, motivated by our preliminary finding that annotation agreement was more strongly associated with the underlying model than with expert persona (Appendix, Figure~\ref{ref:prel-study-pair-kappa}).

\subsubsection{Utterance-Level Disagreement.}\label{sec:utt-level-dis}
For each utterance, we quantify cross-model disagreement using an entropy-based metric, normalized vote entropy over the models' predicted labels~\cite{argamon1999committee}:
\begin{equation}
D(u) =
\frac{-\sum_{c}p_c\log_2 p_c}
{\log_2 n}
\label{eq:entropy}
\end{equation}
where $p_c$ is the proportion of model predictions assigned to label $c$, and $n$ is the number of models. 
We normalize by $\log_2 n$, the maximum possible vote entropy for our model panel, to express disagreement on an interpretable scale from 0 to 1.
Higher entropy indicates more diverse predictions across models and stronger cross-model disagreement, while lower entropy indicates less diverse predictions and greater agreement. 
We define 
$D(u)=0$ when all models predict the same label and $D(u)=1$ when all models predict different labels.

\subsubsection{Class-Level Disagreement}\label{sec:class-level-dis}
We also quantify how strongly cross-model disagreement is associated with each label.
For each label $c$ that receives at least one model vote, we define its disagreement score as the vote-weighted mean of utterance-level entropy:
\begin{equation}
D(c) =
\frac{\sum_{u \in \mathcal{U}_{\mathrm{rev}}} v(c,u)\,D(u)}
{\sum_{u \in \mathcal{U}_{\mathrm{rev}}} v(c,u)}
\label{eq:label-disagreement}
\end{equation}
where $\mathcal{U}_{\mathrm{rev}}$ is the data collection being annotated 
and $v(c,u)$ is the number of models that assign label $c$ to utterance $u$.
An utterance contributes to $D(c)$ only if at least one model assigns it label $c$, and its contribution is weighted by the number of models that do so.
A higher $D(c)$ indicates that models tend to predict $c$ on utterances with stronger cross-model disagreement, while a lower $D(c)$ indicates that they tend to predict $c$ on utterances with greater agreement.
This score measures disagreement associated with a label, not whether the label or any model prediction is incorrect.

Class-level disagreement can help prioritize limited and valuable expert effort.
When reviewing every label is impractical, $D(c)$ can be used to rank or group labels by their approximate difficulty and strategically allocate expert attention.
For example, in our study, we grouped labels into high-, medium-, and low-disagreement tiers.
We then used these tiers to select target labels and distribute expert effort across labels with different disagreement levels.

\subsection{Codebook Verifying (CV): Experts Review AI-Proposed Codebook Revisions\label{sec:CV-method}}
\subsubsection{Interface and Experts' Activities}
\begin{figure*}
    \centering
    \includegraphics[width=0.99\linewidth]{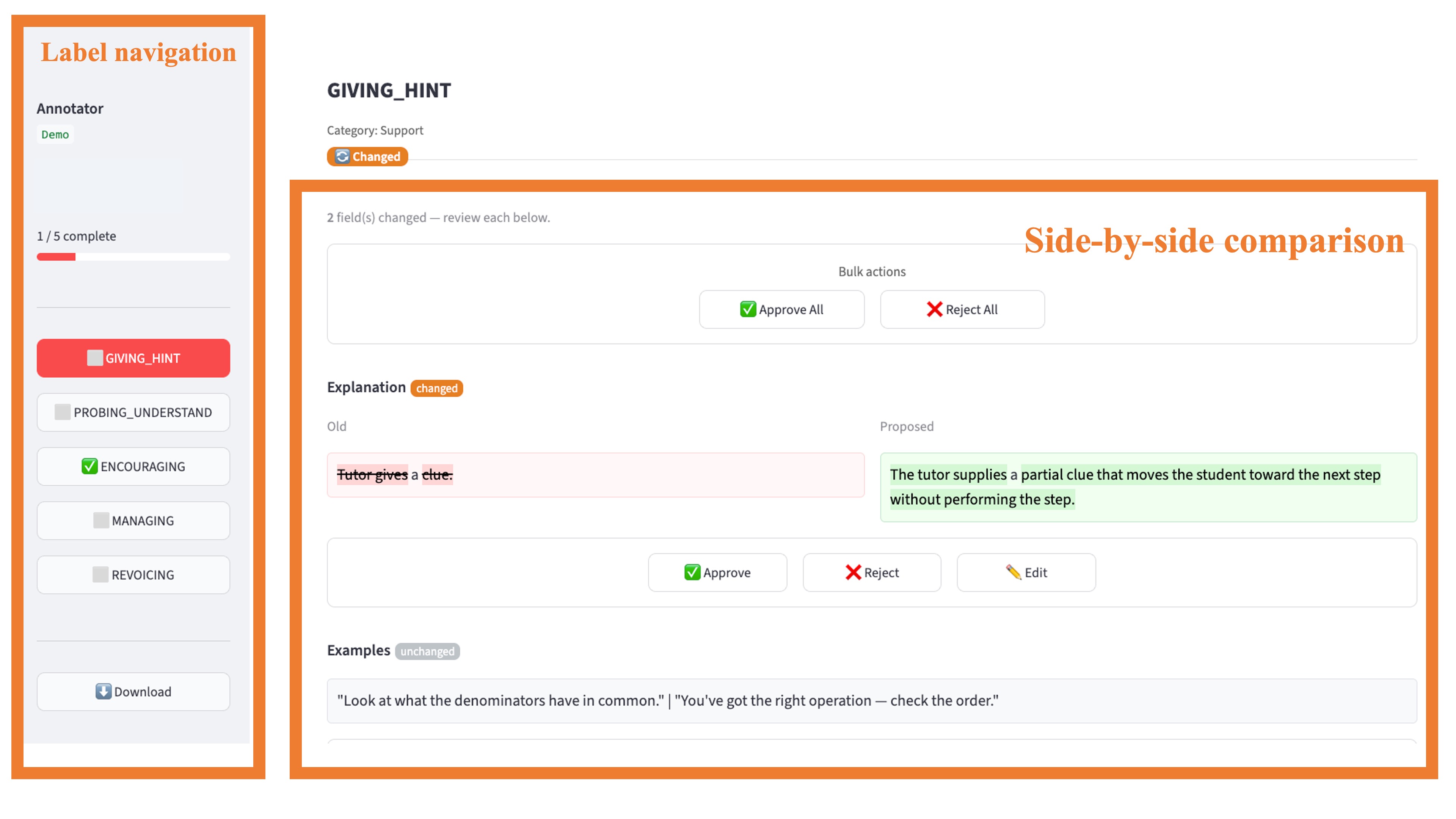}
    \caption{Codebook Verifying Interface. Experts navigate assigned labels using the left sidebar and compare the original codebook entry with LLM-proposed revisions in the main panel. Deletions and additions are highlighted, and experts can approve, reject, or edit each proposed field revision.}
    \label{fig:system-cv}
    \Description{Screenshot of the CV interface. The left sidebar lists assigned labels and review progress. The main panel compares the original and LLM-proposed entries for ``Giving Hint'' side by side, highlighting deletions and additions. Experts can approve, reject, or edit individual revisions, or approve or reject all revisions for the label.}
\end{figure*}

Codebook Verifying (CV) asks experts to review and edit codebook revisions proposed by an LLM.
Figure~\ref{fig:system-cv} shows the interface, which experts use to review the original and LLM-revised entries side by side for each label.
Each label contains five fields that specify how it should be applied: \emph{Explanation}, \emph{Examples}, \emph{Near-Hits}, \emph{Near-Misses}, and \emph{Non-Examples}.
Together, these fields provide general guidance for applying a label to an utterance.
We define each field in detail in Section~\ref{sec:codebook-project}.

For each field, experts can accept the LLM-proposed revision, reject it and retain the original content, or manually edit the proposed content.

\subsubsection{Generating Codebook Revisions for Experts to Review}
Under the hood, we use an LLM to generate candidate revisions for all labels.
For each label, the LLM receives the initial codebook entry and high-disagreement utterances associated with that label, selected from the data and process described in Section~\ref{sec:disagreement-method}.
For each utterance, the LLM also receives the predictions and rationales from all models.
The LLM is then instructed to propose revisions to the five fields of the codebook entry.
These proposed revisions are displayed in the interface (Figure~\ref{fig:system-cv}) for experts to review.
Importantly, the LLM receives no expert guidance on how to resolve the disagreements when generating these revisions.
For example, in a boundary case, the LLM does not know which side of the boundary experts consider correct.
The proposed revisions therefore represent the LLM's best judgment rather than expert-validated changes.

\subsubsection{Forming the Final Codebook}
We do not assume that experts have enough time to review every class label, as discussed in our design constraints (Section~\ref{sec:design-constraint}).
We therefore select a subset of class labels as \emph{target labels} and ask experts to review only those labels.
For labels that experts do not review, we vary whether to retain the initial instructions or use the LLM-revised instructions.
This results in three versions of the final codebook:
{\em (i)} \textbf{Fully LLM-revised}: LLM-revised instructions for all labels without expert validation, serving as the automated revision baseline;
{\em (ii)} \textbf{Expert-validated + initial}: expert-edited instructions for target labels and initial instructions for unreviewed labels; and
{\em (iii)} \textbf{Expert-validated + LLM-revised}: expert-edited instructions for target labels and LLM-revised instructions for unreviewed labels.

\subsection{Question Answering (QA): Experts Answer Questions About Decision Rules\label{sec:QA-method}}
\subsubsection{Interface and Experts' Activities}
\begin{figure*}
    \centering
    \includegraphics[width=0.99\linewidth]{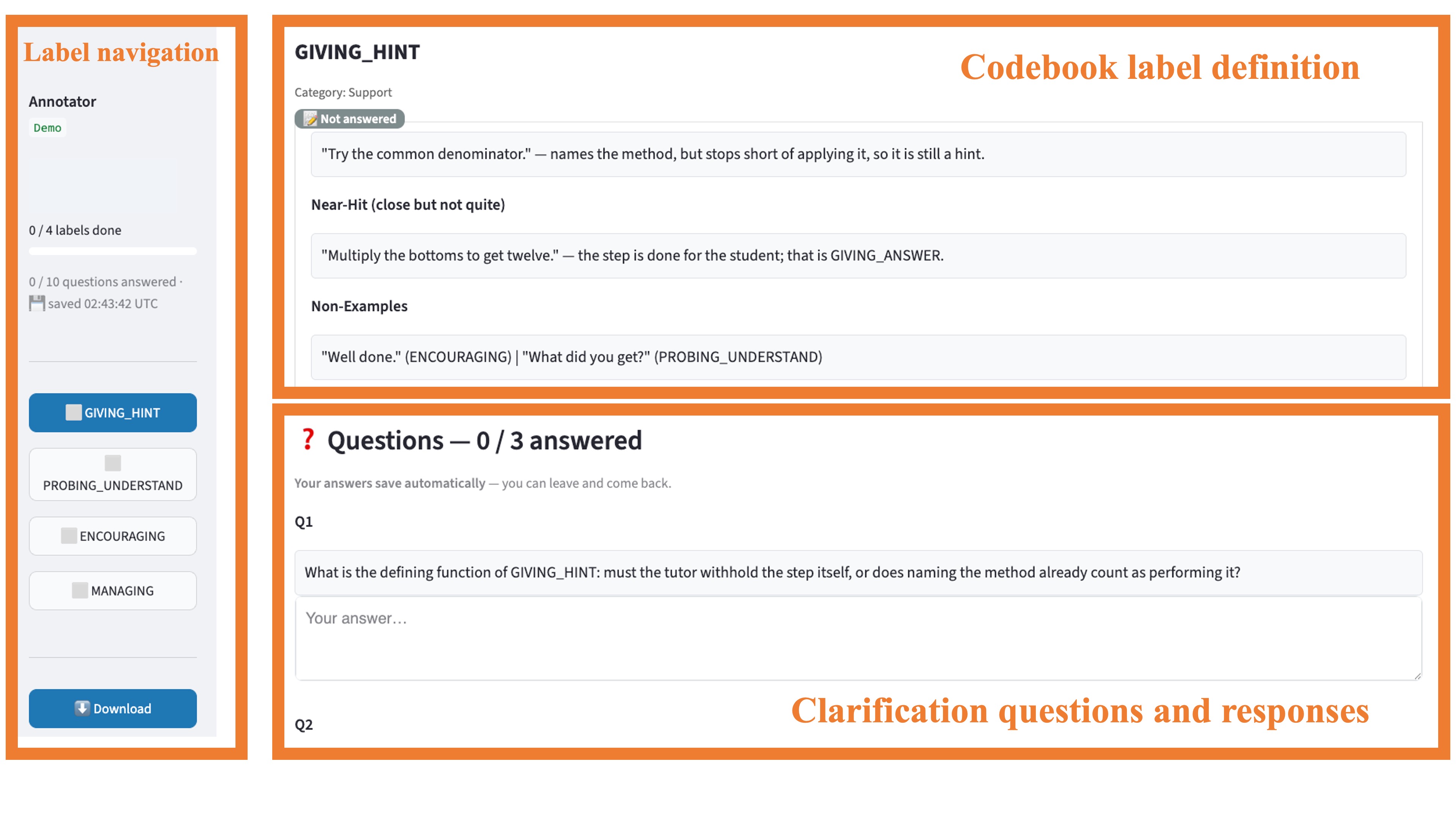}
    \caption{Question Answering Interface. Experts navigate assigned labels using the left sidebar and consult the selected label's codebook entry in the upper panel. The lower panel presents clarification questions generated from ambiguities surfaced by cross-model disagreement. Experts provide written answers that guide subsequent codebook revisions.}
    \label{fig:system-qa}
    \Description{Screenshot of the QA interface. The left sidebar lists assigned labels and completion progress. The upper panel displays the selected ``Giving Hint'' codebook entry. The lower panel presents clarification questions about label ambiguities, with text boxes for experts to provide written answers that guide subsequent codebook revisions.}
\end{figure*}

Question Answering (QA) elicits expert input before an LLM generates the final revised codebook.
Figure~\ref{fig:system-qa} shows the interface, which experts use to answer six questions for each label assigned to them.
An LLM automatically generates these questions from high-disagreement cases to elicit reusable decision rules that can apply to multiple cases with similar patterns, rather than judgments about individual utterances.
Experts answer the questions without seeing the underlying transcripts, model predictions, rationales, or ambiguity analysis.

\subsubsection{Generating Questions About High Disagreement Cases for Experts to Answer\label{sec:qa-method-generation}}
Behind the scenes, our workflow first identifies recurring ambiguity patterns among high-disagreement cases and then generates questions that ask experts how these ambiguities should be resolved.
For each label, an LLM receives the corresponding codebook entry, entries of competing labels, summary statistics of cross-model disagreement, and high-entropy utterances along with the predictions and rationales from multiple LLMs (from the data and process described in Section~\ref{sec:disagreement-method}).
We characterize competing labels using two measures computed over the target label's candidate pool: alternative-choice probability, the proportion of all non-target model votes assigned to a competing label; and co-appearance, the number of contested utterances for which at least one model predicted that competing label. 
Alternative-choice probabilities sum to 100\% across all competing labels.
Using this information, the LLM generates recurring ``ambiguity patterns'', written diagnoses of why the target label is difficult to apply consistently.
The LLM then converts these patterns into six self-contained questions for each label, covering complementary aspects of category specification:

\begin{itemize}
\item \emph{Core behavior}: the tutor action and its function for the student, independent of particular wording;
\item \emph{Necessary condition}: what must be present for the label to apply;
\item \emph{Exclusion}: what resembles the category but should be excluded, and why;
\item \emph{Boundary}: how to distinguish the target from a commonly confused neighboring category;
\item \emph{Scope}: how to apply the label to turns with multiple functions or only a brief instance of the relevant behavior;
\item \emph{Edge case}: how to handle borderline or unusual cases not clearly addressed by the initial entry.
\end{itemize}

These questions elicit reusable decision rules for resolving recurring ambiguities rather than expert judgments on individual utterances.


\subsubsection{Generating the Final Codebook}
We use generated questions and experts' answers to construct four final codebook variants that vary in the evidence provided to the LLM and the labels eligible for revision.
Because 
we do not assume that experts have enough time to answer questions for every label, we select a subset as \emph{target labels} and ask experts to answer questions generated for them.
Although each question is generated for a target label, it may explicitly reference other labels, particularly when asking experts to distinguish between categories.
We therefore vary two aspects of codebook revision: the evidence provided to the LLM and whether the LLM revises only the target labels or also other labels referenced in the questions and answers.
Together, these choices produce four codebook variants:
{\em (i)} \textbf{Disagreement + QA}: the LLM receives the original disagreement cases and corresponding expert answers and revises the target labels;
{\em (ii)} \textbf{QA only}: the LLM receives only the expert answers and revises the target labels;
{\em (iii)} \textbf{Ambiguity Pattern + QA}: the LLM receives the identified ambiguity patterns (Section~\ref{sec:qa-method-generation}) and corresponding expert answers and revises the target labels; and
{\em (iv)} \textbf{Ambiguity Pattern + Within and Cross Label QA}: the LLM receives the ambiguity patterns and expert answers and revises both the target labels and other labels explicitly involved in the expert responses.
In all four variants, labels that are not eligible for revision retain their original codebook instructions.

\subsection{Rationale Labeling (RL): Experts Label and Explain Contested Cases\label{sec:RL-method}}
\subsubsection{Interface and Experts' Activities}
\begin{figure*}
    \centering
    \includegraphics[width=0.99\linewidth]{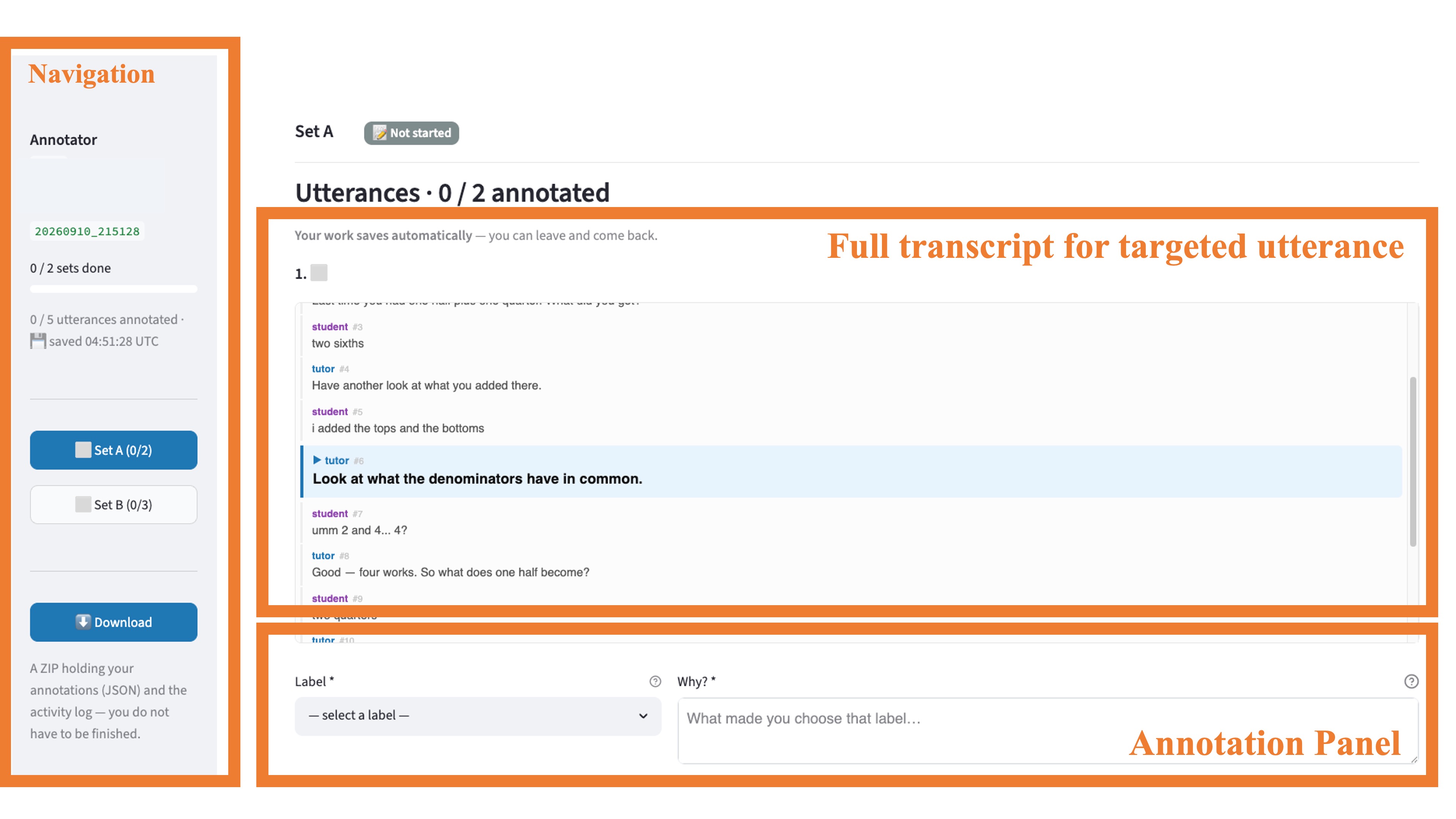}
    \caption{Rationale Labeling Interface. Experts review the highlighted tutor utterance within its surrounding conversation (upper panel), and provide a written rationale (lower panel). The left sidebar supports navigation and tracks annotation progress. These annotations and rationales inform subsequent codebook revisions.}
    \label{fig:system-rl}
    \Description{Screenshot of the RL interface. Experts read a target tutor utterance in its surrounding conversation, assign a codebook label, and provide a written rationale explaining their decision. These annotations and rationales guide subsequent codebook revisions.}
\end{figure*}

Rationale Labeling (RL) elicits expert judgments on individual disagreement cases before an LLM generates the final revised codebook.
Figure~\ref{fig:system-rl} shows the interface, which experts use to review each utterance, select the label they believe best applies, and provide a free-text rationale.
Both the label and rationale are required.
Experts can select any label in the codebook.

Unlike in the other two approaches, experts in RL do not see the target label that motivated the selection of each utterance.
Instead, utterances selected across all target labels are pooled and randomly shuffled before being presented to the expert.
RL hides the target label to avoid bias.


\subsubsection{Selecting Disagreement Cases for Experts to Label}
Under the hood, we use a subset of \emph{target labels} to prioritize expert effort.
For each target label, we first identify utterances from the multi-LLM annotated data described in Section~\ref{sec:disagreement-method} for which at least one model predicts the target label.
From this pool, we select 10 utterances per target label using a greedy procedure. 
At each step, we select the utterance that introduces the most competing labels not yet represented in the selected set, breaking ties by prioritizing higher entropy scores.
This procedure promotes coverage of diverse competing labels while limiting repeated examples of the same label distinction.
Each utterance is assigned to only one target label across the study.
We process target labels in ascending order of eligible pool size to preserve available examples for labels with fewer candidates. 
This coverage reflects model predictions, not expert-confirmed labels.

The target labels determine which utterances enter the annotation task but are hidden from experts.
Experts also do not see any model predictions. 
This allows the target labels to guide the allocation of expert effort without biasing experts toward particular labels when making their judgments.

\subsubsection{Generating the Final Codebook}
Rationale Labeling produces one final codebook, \textbf{RL-revised}.
An LLM uses the expert-assigned labels and rationales to revise the codebook entries for all labels represented in the expert annotations, while retaining the initial entries for all remaining labels.
Although utterances are selected using target labels, experts can assign any label to each utterance.
As a result, expert input tends to spread across non-target labels more broadly than in Question Answering.

\section{Study Context, Design, and Setup}

In this paper, we evaluated the proposed workflow (Section~\ref{sec:method}), including its three approaches to eliciting expert input, within a large-scale tutoring-session transcript annotation project.
The project aimed to understand what instructional actions and teaching strategies tutors use in large-scale, real-world data.
This section describes the annotation project context (Section~\ref{sec:annotation-project}), study design and setup (Section~\ref{sec:study-design}), as well as implementation of the workflow in our study (Section~\ref{sec:implementation}).

\subsection{Real-World Tutoring Session Transcript Annotation Project\label{sec:annotation-project}}

Our study was embedded in an independently motivated codebook development project aiming to understand real-world data.
Over six months, our research team worked with two education experts to develop and iteratively refine a taxonomy of tutoring moves that characterizes potentially valuable instructional actions and supports fine-grained annotation of tutoring discourse. 
This process was conducted independently of the present study rather than constructed for experimental purposes.
As experts annotated new tutoring interactions, resolved disagreements, and refined their coding criteria, the project preserved successive snapshots of the codebook together with the expert-consensus annotations produced during each stage.

\subsubsection{Data Sources, Licensing, and Processing\label{sec:tutor-data-context}}
The data for this study consisted of transcripts of mathematics tutoring sessions from three remote human tutoring providers serving students in Grades 3 to 12 (around ages 9 to 18) in the United States and United Kingdom.
The three providers covered substantially different tutoring formats:

\begin{itemize}
    \item 
\textbf{Eedi} is a U.K. online learning platform for students aged 9--16. 
Students complete problem sets assigned by their teachers and can optionally initiate a text-based chat with a human tutor when they need assistance with a particular problem ~\cite{zent2025piivot}.
These interactions typically produce relatively short, problem-focused student-tutor conversations. 
On average, there are 34 utterances per tutoring session for this provider.  

\item
\textbf{UPChieve} is a nonprofit online tutoring platform providing free, on-demand, one-to-one academic support primarily to U.S. middle- and high-school students. 
Students are matched with trained volunteer tutors and communicate within a virtual classroom supporting text chat, a shared whiteboard or document editor, and optional voice communication.
On average, there are 116 utterances per tutoring session for this provider.  

\item
\textbf{Third Space Learning} provides online tutoring to grade-school students through longer, structured lessons organized around predetermined instructional materials and slides. Unlike the primarily text-based interactions of the other two platforms, tutoring on TSL is conducted through spoken conversation between a student and a human tutor, with the resulting audio subsequently transcribed into text.
On average, there are 500 utterances per tutoring session for this provider. 

\end{itemize}

The data were shared with the authors of this paper in accordance with the providers' user agreements,\footnote{We do not report the total number of sessions because the dataset is continuously growing.} and the study was conducted under approval from the researchers' Institutional Review Board. 
The providers transcribed all audio data using automated speech recognition before sharing them with the research team.
All data were de-identified before annotation to ensure that annotators were not exposed to personally identifiable information.


\subsubsection{Experts\label{sec:experts}}
This annotation project involved two education experts who developed and maintained the codebook through an iterative annotation process.
Both are veteran mathematics educators with extensive experience teaching in individual, group, and classroom settings.
Both hold doctoral degrees in Learning Sciences and Technologies, are active researchers, and have prior experience qualitatively coding educational data.



\subsubsection{Expert Codebook Development Process and Outcomes\label{sec:codebook-project}}
The two experts developed the codebook through weekly annotation and discussion.
Each week, they independently annotated new tutoring utterances, compared their judgments, discussed disagreements until reaching consensus, and revised the codebook to clarify unclear or underspecified distinctions.
Once they reached consensus on an utterance, they finalized its label and did not revisit it as the codebook evolved.

\paragraph{Labels.}
To characterize tutor actions, the research team developed a taxonomy based on prior research in cognitive science, learning sciences, classroom discourse analysis, and intelligent tutoring systems~\cite{zhou2026tutor}.
The expert-developed taxonomy contains 29 tutoring move labels.
During annotation, experts left an utterance unlabeled if none of the 29 tutoring moves applied.
For this paper's classification tasks, we treat these cases as an additional \textsc{None} class, resulting in 30 possible labels for each tutor utterance.

\paragraph{Codebook Structure.}
Each of the 29 tutoring move labels is defined through five fields: an \emph{Explanation} defining the label's scope; \emph{Examples} showing clear instances; \emph{Near Hits} showing similar cases that do not qualify; \emph{Near Misses} showing borderline cases that still qualify; and \emph{Non Examples} showing cases outside the category.

In the initial codebook, every label included an Explanation, but the other four fields were often sparse.
As experts encountered new cases during annotation, they expanded these fields to clarify how each label should be applied.

\paragraph{Codebook Snapshots.}

This process produced two codebook snapshots:
\begin{itemize}
\item \textbf{Initial codebook (January 2026).} The earliest expert-developed version of the codebook and the starting point for the revision methods studied in this paper. The experts produced 825 finalized tutor-utterance annotations during its development.


\item \textbf{Final codebook (July 2026).} Following the initial snapshot, the experts continued applying and refining the codebook while producing an additional 4,032 finalized tutor-utterance annotations. At the conclusion of this process, the experts conducted a final review of the codebook, producing the final expert-reviewed version used in this study. The 4,032 annotations produced during this period serve as the held-out evaluation set for our main study.
After completing the Final Codebook, the two experts stopped annotating data and revising the codebook.
\end{itemize}

Once the experts reached consensus on an utterance, its annotation was finalized and was not updated as the codebook evolved.
When the experts could not resolve a disagreement, they tabled the case and revisited it later, after further codebook development provided clearer guidance.
Throughout the project, the experts, together with the paper authors, periodically returned to these unresolved cases and used the latest codebook to reach consensus.
Thus, the final annotation set preserves previously agreed judgments while incorporating later resolutions of cases that were initially unresolved.
The final codebook reflects the experts' accumulated understanding of the taxonomy after the full annotation and revision process.

\subsection{Study Design and Setup\label{sec:study-design}}


The longitudinal record of this annotation project provides a unique opportunity to evaluate our proposed workflows. 
In our study, each workflow revised the same Initial Codebook (Section~\ref{sec:codebook-project}) using the same previously unseen tutoring data.
Section~\ref{sec:revision-eval-data} describes the data used for revision and evaluation.

Because expert time is limited, each expert provided input for only a subset of target labels in each workflow.
We assigned target labels to balance label difficulty 
across conditions while minimizing learning effects across workflows.
Each expert first completed CV, followed by QA and RL, with one to two weeks between conditions.
Section~\ref{sec:target-label-assign} describes the target label assignment procedure in detail.

Finally, we evaluated the resulting codebooks by measuring LLM labeling performance on the held-out evaluation test set.
Sections~\ref{sec:multi-llm-annotation} and~\ref{sec:eval-setups} describe the LLM annotation and evaluation procedure.

\subsubsection{Data Used\label{sec:revision-eval-data}}
To prevent data leakage between codebook revision and evaluation, we kept the data used for LLM codebook revision separate from the expert-annotated data used for evaluation.
The LLM codebook revision pipeline had no access to either the sessions or expert labels in the evaluation set.
We used three separate datasets in our experiments:

\begin{itemize}

    \item \textbf{Heldout Evaluation Test Set (110 Sessions; 4,032 Expert Labels).}
    As the experts developed the Initial Codebook into the Final Codebook, they manually annotated tutoring data, discussed disagreements, and reached consensus on 4,032 tutor utterances from 110 sessions.
    Within the workflow study, these sessions and labels were used only for evaluation and were not supplied to the automated revision pipeline. They originated in the earlier longitudinal annotation process that also informed the Final Codebook and the participating experts' understanding of the task.

    \item \textbf{Unlabeled Data for LLM Codebook Revision (165 Sessions; 6,595 Unlabeled Utterances).}
    We separately sampled 165 sessions containing 6,595 tutor utterances to form the unlabeled revision corpus used by all three LLM codebook revision workflows.
    This corpus contains 100 sessions and 2,245 tutor utterances from Eedi, 50 sessions and 2,277 tutor utterances from UPChieve, and 15 sessions and 2,073 tutor utterances from TSL.
    Because session length varies substantially across providers, we sampled different numbers of sessions to approximately balance the number of tutor utterances contributed by each provider.
    This corpus was used for multi-LLM annotation and the calculation of cross-model disagreement described in Section~\ref{sec:disagreement-method}.

    \item \textbf{Early Expert Annotation Set (14 Sessions; 825 Expert Labels).}
    While developing the Initial Codebook, the experts produced 825 finalized tutor utterance annotations from 14 sessions.
    We kept these annotations separate from both the revision corpus and the held-out evaluation test set.
    We used them only for additional analyses, including early exploratory analyses and our analysis of cross-model disagreement in Section~\ref{sec:findings-rq-3}.

\end{itemize}

\subsubsection{Experimental Setup: Assigning Target Labels Across Workflows\label{sec:target-label-assign}}


Expert time and effort are limited and valuable.
Rather than asking experts to review the entire codebook under each workflow, we strategically allocated their effort to a subset of labels, called \textit{target labels} (Section~\ref{sec:method}).
\textbf{Each expert was assigned six target labels per workflow}, for a total of 12 target labels in each workflow.
In CV and QA, the target labels determined which codebook entries or questions experts reviewed.
In RL, they determined which pools were used to select cases for expert annotation.

\paragraph{Target Label Assignment for CV}
For CV, we divided the labels into high, medium, and low disagreement groups based on their cross-LLM disagreement ranking (Section~\ref{sec:class-level-dis}).
We selected the four labels with the highest disagreement from the high group, the middle four labels from the medium group, and the four labels with the lowest disagreement from the low group, yielding 12 target labels.
This selection covered a broad range of disagreement levels.
We divided the 12 labels evenly between the two experts.

\paragraph{Target Label Assignment for QA and RL}
For QA and RL, we selected another 12 labels immediately adjacent to the CV labels in the disagreement ranking: four immediately below the CV labels in the high-disagreement group, two immediately above and two immediately below the CV labels in the medium-disagreement group, and four immediately above the CV labels in the low-disagreement group.
This gave QA and RL labels at similar disagreement levels without overlapping the CV target labels.
We divided these 12 labels between the two experts and swapped their assignments between QA and RL.
Thus, neither expert worked on the same target label in both workflows, allowing us to compare QA and RL while reducing potential learning effects.
Some carryover remained possible despite our efforts to minimize it through the study design, because QA questions could reference other labels and RL allowed experts to assign any label.
This carryover is inherent to evaluating all three workflows with the same coding scheme and cannot be practically eliminated.\footnote{Before the main study, we conducted a small pilot using an earlier version of CV. Because this pilot preceded the complete revision corpus, we calculated its disagreement ranking using a different data sample. A small number of pilot labels therefore overlap with the later QA and RL target labels.
We report the pilot procedure and overlap in Appendix~\ref{app:disagreement-entropy-ranking-assignment}.}

\paragraph{Expert Workload}
For the expert workload, we used six questions per target label in QA and 10 utterances per target label in RL.
Each expert reviewed codebook revisions for six labels in CV, answered 36 questions in QA (6 labels $\times$ 6 questions), and annotated 60 utterances with rationales in RL (6 labels $\times$ 10 utterances).
The amount of time this took averaged 53.1 minutes (0.9 hours) per expert for CV, 179.2 minutes (3.0 hours) for QA, and 169.0 minutes (2.8 hours) for RL.

\subsubsection{Multi-LLM Data Annotation\label{sec:multi-llm-annotation}}

A key component of our workflow is using multiple LLMs to independently annotate the same data with a given codebook.
We used this process for two purposes:

\begin{itemize}
    \item \textbf{Measuring cross-LLM disagreement for LLM codebook revision.}
    We used the Initial Codebook to annotate all 6,595 tutor utterances in the unlabeled revision corpus (Section~\ref{sec:revision-eval-data}) with six LLMs from different providers: OpenAI's GPT 5.1, Google's Gemini 2.5 Pro, Anthropic's Claude Sonnet 4.5, Alibaba's Qwen3 Coder 480B A35B, DeepSeek V3.2, and Mistral Large 3 675B Instruct.\footnote{The exact model identifiers were \nolinkurl{openai.gpt-5.1.2025-11-13}, \nolinkurl{google.gemini-2.5-pro}, \nolinkurl{anthropic.claude-4.5-sonnet}, \nolinkurl{qwen.qwen3-coder-480b-a35b}, \nolinkurl{deepseek.v3.2}, and \nolinkurl{mistral.mistral-large-3-675b-instruct}, respectively.}
    Each model produced a predicted label and a free-text rationale for each utterance.
    We used these predictions to calculate utterance- and class-level disagreement as described in Section~\ref{sec:disagreement-method}.

    \item \textbf{Producing final annotations for codebook evaluation.}
    To evaluate each resulting codebook, we used the same six LLMs to independently annotate the held-out test set of 4,032 expert consensus annotations (Section~\ref{sec:revision-eval-data}).
    We aggregated the six predicted labels using eight methods implemented in Crowd Kit~\cite{CrowdKit}: Majority Vote, Dawid Skene, One Coin Dawid Skene, GLAD, MMSR, Wawa, Zero Based Skill, and MACE.
    We used multiple LLMs and aggregated their predictions because prior work has shown that combining multiple annotators can produce more robust annotation outcomes than relying on a single annotator~\cite{10.1145/3613904.3642834}.
\end{itemize}

\subsubsection{Codebook Evaluation\label{sec:eval-setups}}
For each codebook and aggregation method, we computed accuracy and weighted F1 against the recorded expert-consensus labels on the 4,032-utterance evaluation set. The main results represent arithmetic means across the eight aggregation methods, not the performance of a single combined predictor. Each annotation model completed one run per codebook at temperature 0, without persona conditioning. The aggregation methods provide alternative summaries of the same model predictions rather than independent experimental replications.
For the 6.1\% of utterances with multiple expert consensus labels, we used the first recorded label as the single-label reference.
These metrics therefore measure agreement with one recorded label per utterance, rather than whether a prediction matches any expert-endorsed label.

All revision workflows started from the same Initial Codebook, drew disagreement evidence from the same unlabeled revision corpus, used the same LLMs for annotation, and were evaluated on the same held-out test set.
These shared components provide a common basis for comparing the resulting codebooks.
However, the workflows tested in our study differed in target label assignments, expert exposure, and required expert effort.
Our study therefore compares the overall revision strategies rather than isolating the causal effect of the form of expert input alone.

\subsection{Workflow Implementation\label{sec:implementation}}

\subsubsection{Disagreement Calculation}
For each utterance and class, we computed cross-LLM disagreement from the six LLM predictions (Section~\ref{sec:multi-llm-annotation}) using the method described in Section~\ref{sec:disagreement-method}.
We then ranked labels by $D(c)$ from highest to lowest and used this ranking to assign target labels across workflows (Section~\ref{sec:target-label-assign}).
The complete ranking and assignments are reported in Appendix~\ref{app:disagreement-entropy-ranking-assignment}.

\subsubsection{Workflow Implementation}

For CV, GPT 5.6 Sol generated candidate revisions for all 30 labels, and experts reviewed the 12 target labels, six per expert (Section~\ref{sec:target-label-assign}). 
For QA, GPT 5.6 Sol generated six questions for each target label, and each expert answered 36 questions across six labels, yielding 72 responses that GPT 5.6 Sol used to revise the codebook (Section~\ref{sec:QA-method}). 
For RL, each expert labeled and provided rationales for 60 selected disagreement utterances, 10 for each target label, yielding 120 annotations in total (Section~\ref{sec:RL-method}). Experts assigned 24 distinct labels across these cases, which GPT 5.6 Sol used to revise the corresponding 24 codebook entries; the remaining six retained their original text.
Appendix~\ref{app:prompts-experimental-details} provides detailed prompts and evidence configurations.

\section{Findings}
\subsection{RQ1:~Annotation Performance of the Revision Workflows\label{sec:findings-rq-1}}

\begin{table*}[t]
\centering

\small
\setlength{\tabcolsep}{7pt}
\renewcommand{\arraystretch}{1.12}

\begin{tabular}{@{}llrrr@{}}
\toprule
\textbf{Workflow} &
\textbf{Codebook condition} &
\textbf{Labels eligible for revision} &
\textbf{Accuracy ($\uparrow$)}  &
\textbf{Weighted F1 ($\uparrow$)} \\
\midrule

Expert-only
& Initial codebook
& 0 & 0.534& 0.536\\

\midrule

AI-only
& Fully LLM-revised
& 30 & 0.531
& 0.547\\

\midrule

CV
& Expert-validated + initial& 12 & 0.520& 0.521\\

& Expert-validated + LLM-revised
& 30 & 0.537
& 0.557\\

\midrule

QA
& Disagreement + QA
& 12 & 0.552& 0.565\\

& QA-only
& 12 & 0.562
& 0.567\\

& Ambiguity-pattern + QA
& 12 & 0.580
& 0.586\\

& Ambiguity Pattern + Within and Cross Label QA
& 25 & \underline{0.605}
& \underline{0.613}\\

\midrule

RL
& RL-revised
& 24 & \textbf{0.649}
& \textbf{0.658}\\
\midrule
 Expert-only& Final codebook& 30& 0.578&0.583\\
\bottomrule

\end{tabular}
\caption{
Performance of the expert-only and revised codebooks on the 4,032-utterance evaluation set, including \textsc{None}.
Accuracy and weighted F1 are averaged across eight aggregation methods;
higher values indicate better performance. 
RL achieves the highest accuracy and weighted F1, followed by Ambiguity Pattern + Within and Cross Label QA; both outperform the final expert-developed codebook on both metrics.
``Labels eligible for revision'' denotes the number of codebook labels eligible for change under each condition.
The final codebook was produced through the longitudinal development process. 
\textbf{Bold} and \underline{underlined} values indicate the best and second-best performance, respectively, in each metric column.}
\label{tab:codebook-performance-all}
\Description{Table comparing codebook conditions, labels eligible for revision, and mean accuracy and weighted F1 across eight aggregation methods. RL achieves the highest accuracy (0.649) and weighted F1 (0.658), followed by Ambiguity Pattern + Within and Cross Label QA (0.605 and 0.613). Both outperform the final expert-developed codebook (0.578 and 0.583).}
\end{table*}

\paragraph{Rationale Labeling (RL) performed best and outperformed the expert-revised codebook.}
Table~\ref{tab:codebook-performance-all} compares annotation performance using the Initial Codebook, the three proposed workflows, and the Final Codebook that experts produced.
Accuracy and weighted F1 are averaged across the eight label aggregation methods described in Section~\ref{sec:multi-llm-annotation}.
\textbf{RL achieved the highest performance, with 64.9\% accuracy and 0.658 weighted F1, substantially outperforming the Final Codebook}, which was revised by experts, at 57.8\% accuracy and 0.583 weighted F1.
The best QA condition also outperformed the Final Codebook, reaching 60.5\% accuracy and 0.613 weighted F1.

\paragraph{Question-Answering (QA) generally improved performance, especially when using ambiguity patterns and allowing revision beyond the target labels.}
Table~\ref{tab:codebook-performance-all} also shows that QA generally improved over the Initial Codebook in both accuracy and weighted F1.
Providing ambiguity patterns alongside expert questions and answers was more effective than providing the original disagreement cases, suggesting that abstracting disagreement into recurring patterns may help codebook revision.
Performance was also higher when the LLM could revise labels referenced in expert responses beyond the 12 target labels, rather than being restricted to the target labels.

\paragraph{AI-only revision and Codebook Verifying (CV) did not consistently improve over the Initial Codebook.}
Table~\ref{tab:codebook-performance-all} also shows that the AI-only condition did not consistently improve over the Initial Codebook despite revising all 30 labels.
We hypothesize that, for boundary cases requiring a choice between competing labels, LLMs can only make an educated guess without expert guidance.
CV results were also mixed.
CV performed worse than the Initial Codebook when non-target labels retained their original entries, but slightly better when they used LLM-revised entries.


\begin{table*}[t]
\centering
\small
\setlength{\tabcolsep}{7pt}
\begin{tabular}{llrrrr}
\toprule
\textbf{Workflow} & \textbf{Codebook condition}
& \textbf{Overall}
& \textbf{High entropy}
& \textbf{Mid entropy}
& \textbf{Low entropy} \\
& & $(n=4{,}032)$ & $(n=1{,}141)$
& $(n=740)$ & $(n=2{,}151)$ \\
\midrule

Expert-only
& Initial codebook
& 0.536& 0.295& 0.546& 0.660
\\

\midrule

CV
& Fully LLM-revised
& 0.547& 0.323& 0.519& 0.674
\\
& Expert-validated + Initial& 0.521& 0.263& 0.535& 0.655
\\
& Expert-validated + LLM-revised
& 0.557& 0.346& 0.517& \underline{0.678}
\\
\midrule

QA
& Disagreement + QA
& 0.565& 0.356& \underline{0.591}& 0.664
\\
& QA only
& 0.567& 0.391& 0.550& 0.668
\\
& Ambiguity patterns + QA
& 0.586& 0.432& 0.567& 0.676
\\
& Ambiguity patterns + Within and Cross label QA
& \underline{0.613}& \underline{0.526}& 0.570& 0.674
\\

\midrule

RL
& Annotation + Rationale
& \textbf{0.658}& \textbf{0.568}& \textbf{0.595}& \textbf{0.727}
\\
\midrule

Expert-only & Final codebook& 0.583& 0.429& 0.581&0.667\\

 \bottomrule
\end{tabular}
\caption{
Annotation \textbf{weighted-F1} overall and across high-, mid-, and
low-entropy label groups. Each group contains 10 labels;
$n$ denotes the number of evaluation utterances.
Bold indicates the highest value in each column.
}
\Description{Table comparing weighted F1 across codebook conditions overall and within high-, medium-, and low-entropy groups, each containing 10 labels. RL performs best in all three groups, scoring 0.568, 0.595, and 0.727, respectively. Across all conditions, performance is lowest for high-entropy labels and highest for low-entropy labels.}
\label{tab:accuracy-by-entropy}
\end{table*}

\paragraph{Rationale Labeling (RL) and Question-Answering (QA) substantially improved performance for high disagreement labels.}
Table~\ref{tab:accuracy-by-entropy} breaks down weighted F1 by labels with high, medium, and low disagreement.
It shows that RL substantially improved performance for high-disagreement labels, with weighted F1 increasing from 0.295 to 0.568.
The best QA condition also improved these labels to 0.526.
CV showed mixed results for high-disagreement labels: all conditions improved performance for high-disagreement labels except the CV variant that combined expert-validated revision with initial entries for unreviewed labels. All CV variants reduced weighted F1 for medium-disagreement labels.
Table~\ref{tab:accuracy-by-entropy} also shows that, across all conditions, low-disagreement labels achieved the highest performance, while high-disagreement labels achieved the lowest.
This supports our assumption that cross-LLM disagreement is a useful indicator of annotation difficulty.



\begin{figure*}[t]
    \centering
    \includegraphics[width=0.99\linewidth]{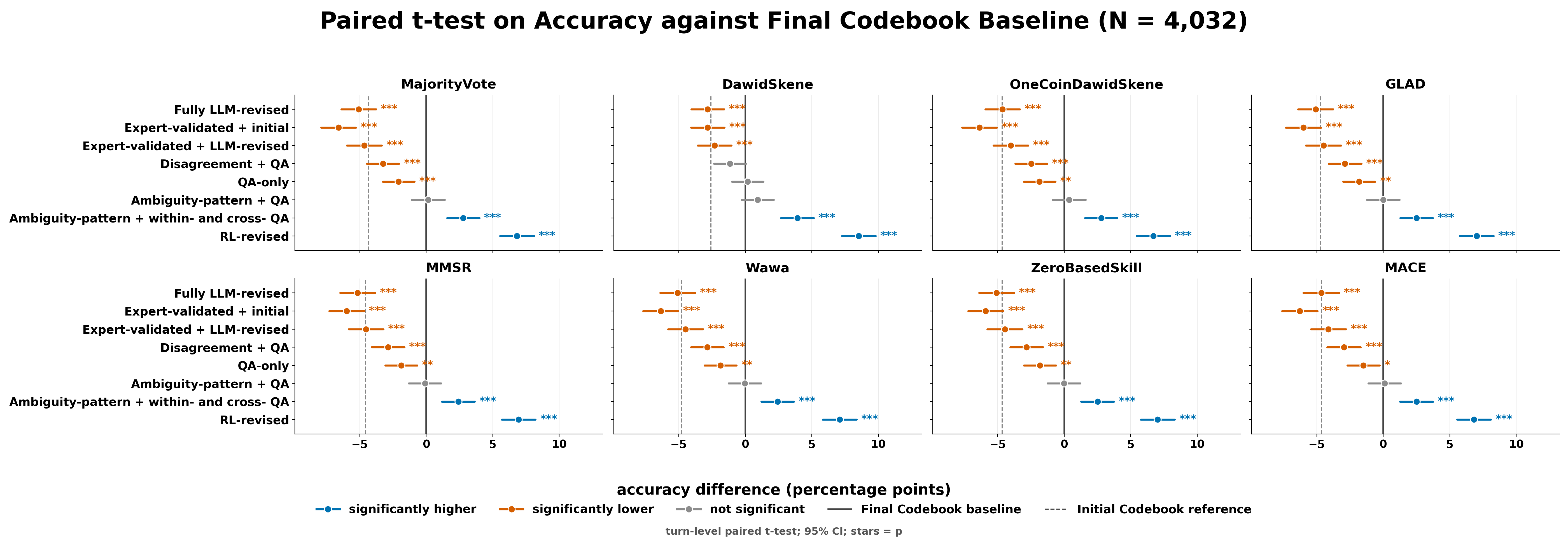}
    \caption{Paired t-test results comparing each revised codebook's accuracy with the final codebook across eight aggregation methods on 4,032 evaluation utterances. Points show accuracy differences in percentage points; error bars indicate 95\% confidence intervals. Solid vertical lines mark equal accuracy, and dashed lines indicate the initial codebook's relative performance. Blue, orange, and gray indicate significantly higher, significantly lower, and nonsignificantly different accuracy, respectively. \textit{RL} and \textit{Ambiguity pattern + Within and Cross label QA} significantly outperform the final codebook across all eight methods, with RL consistently achieving the largest gain.}
    \Description{Eight-panel plot showing paired t-test comparisons with the final codebook on 4,032 evaluation utterances, with one panel per aggregation method. Points show accuracy differences in percentage points, with 95\% confidence intervals; zero indicates equal accuracy. RL and Ambiguity Pattern + Within and Cross Label QA significantly outperform the final codebook across all eight methods, with RL consistently achieving the largest gain.
}
    \label{fig:main-t-test-all}
\end{figure*}

\paragraph{RL consistently and significantly outperformed expert revision, while the best QA condition also significantly outperformed it.}
Figure~\ref{fig:main-t-test-all} shows paired t-test results comparing each revised codebook with the Final Codebook across all eight label aggregation methods.
A difference of zero indicates equal accuracy.
RL significantly outperformed the Final Codebook across all eight methods.
The best QA condition also significantly outperformed it across all eight methods, although by a smaller margin.
Figure~\ref{fig:main-t-test-all} also shows that aggregation method had little effect on the stronger conditions, while weaker conditions such as AI-only revision were more sensitive to the choice of aggregation method.

\subsection{RQ2:~Expert Effort and Experience\label{sec:findings-rq-2}}

\begin{figure*}
    \centering
    \includegraphics[width=0.99\linewidth]{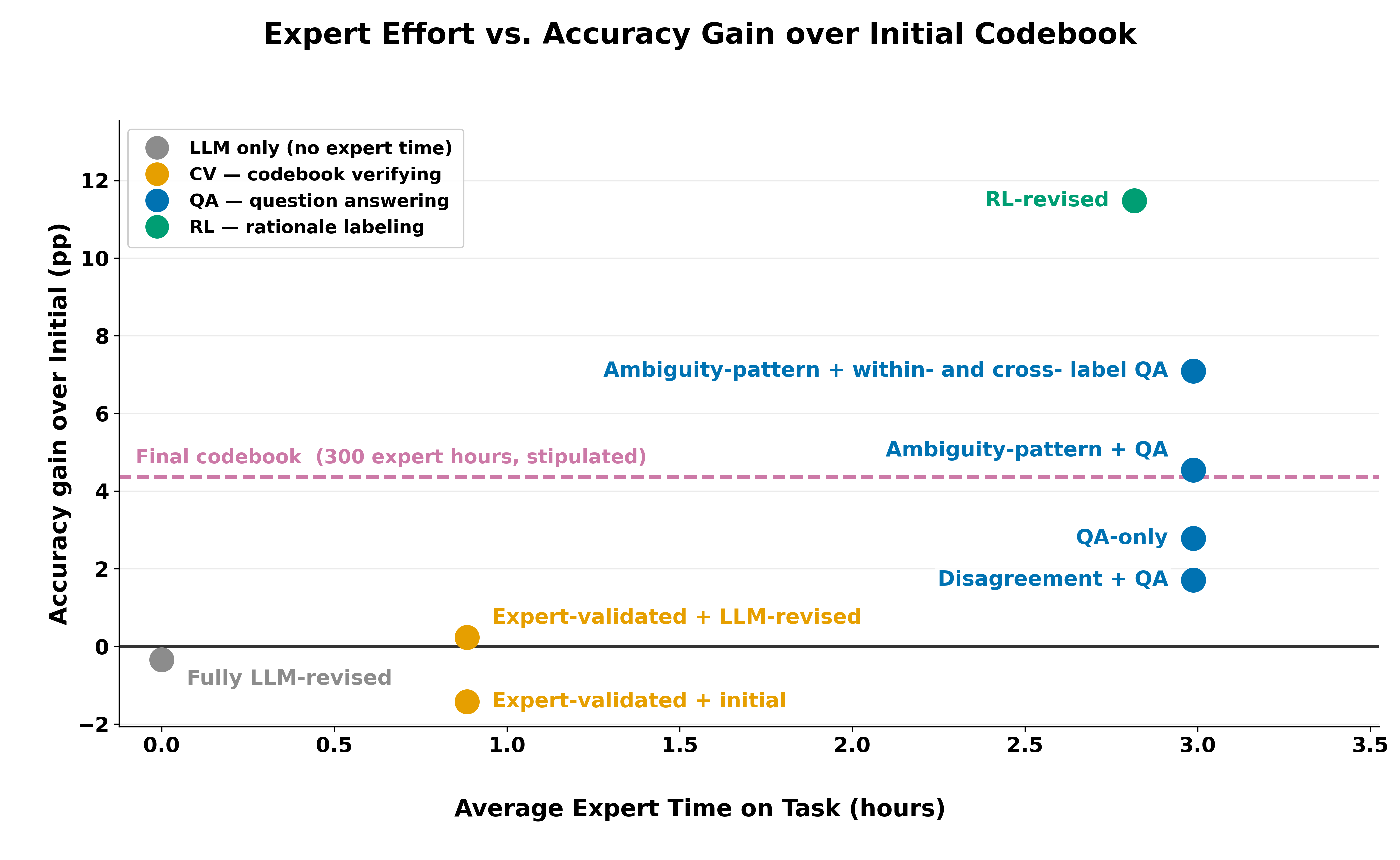}
    \caption{Expert effort and annotation performance gains relative to the Initial codebook. The x-axis shows average expert time on task; the y-axis shows gains in accuracy percentage points. Each point represents a codebook condition, colored by revision workflow. Dashed lines indicate the gains achieved by the Final codebook. RL achieved the largest gains, with slightly less expert time than QA.}
    \label{fig:effort-accuracy}
    \Description{Scatter plot of average expert time in hours versus accuracy gain over the initial codebook in percentage points. RL achieves the largest gain, approximately 11.5 points in 2.8 hours, using slightly less time than QA. CV requires less time but yields little improvement or decreased accuracy. A dashed horizontal line marks the final codebook's approximately 4.4-point gain.
}
\end{figure*}

\paragraph{Rationale Labeling (RL) achieved the largest accuracy gain with only a few hours of expert effort.}
Figure~\ref{fig:effort-accuracy} compares expert time with accuracy gains over the Initial Codebook.
RL required about three hours per expert while achieving the largest accuracy gain, making it the best workflow when annotation performance is the priority.
In contrast, CV required expert time but provided little or no improvement over AI-only revision.
Prior work has also found mixed results when experts manually revise LLM instructions for large-scale annotation, even through iterative revision~\cite{he2025promptingdark}.
Together, these results suggest that asking experts to manually verify and edit LLM-proposed codebook revisions may not be an effective use of limited expert effort.

For reference, Figure~\ref{fig:effort-accuracy} also shows the gain from the Final Codebook. Each expert retrospectively estimated spending approximately 300 hours of work on the project. This estimate concerns the broader annotation and codebook-development process, whereas the workflow timings record additional interaction with the revision tasks. We report it as context, rather than as a directly comparable estimate of the time needed to achieve the observed performance improvement.

We also measured expert effort by the amount of writing required in each workflow.
Figure~\ref{fig:word-vs-acc} in Appendix~\ref{app:additional-analysis-effort} shows a similar trend between expert writing and accuracy gains.


\paragraph{QA offered a useful tradeoff by producing questions and ambiguity patterns alongside performance gains.}
Figure~\ref{fig:effort-accuracy} also shows that QA required similar expert time as RL but produced smaller accuracy gains.
If annotation accuracy is the primary goal, RL is the better choice.
However, QA offers a potentially useful tradeoff because it also produces questions and ambiguity patterns as byproducts.
Both the authors of the paper and the experts found these questions challenging and thought-provoking. 
As is shown in Appendix~\ref{app:survey}, \textbf{QA was rated by the two participating experts as the highest in required mental effort with a score of 7 out of 7}, compared to 5 out of 7 for CV and 3.5 out of 7 for RL.
Meanwhile, QA was also rated the highest among the three in eliciting important label distinctions that were previously unconsidered (QA: 7/7, RL: 5.5/7, and CV: 5/7).  
This suggests that these questions could be used to support broader codebook development.
This may be particularly useful early in the process, when data are still limited, and such questions can help experts identify and discuss important ambiguities.






\subsection{RQ3:~Cross-Model Disagreement and Annotation Difficulty\label{sec:findings-rq-3}}
\begin{figure}
    \centering
    \includegraphics[width=0.8\linewidth]{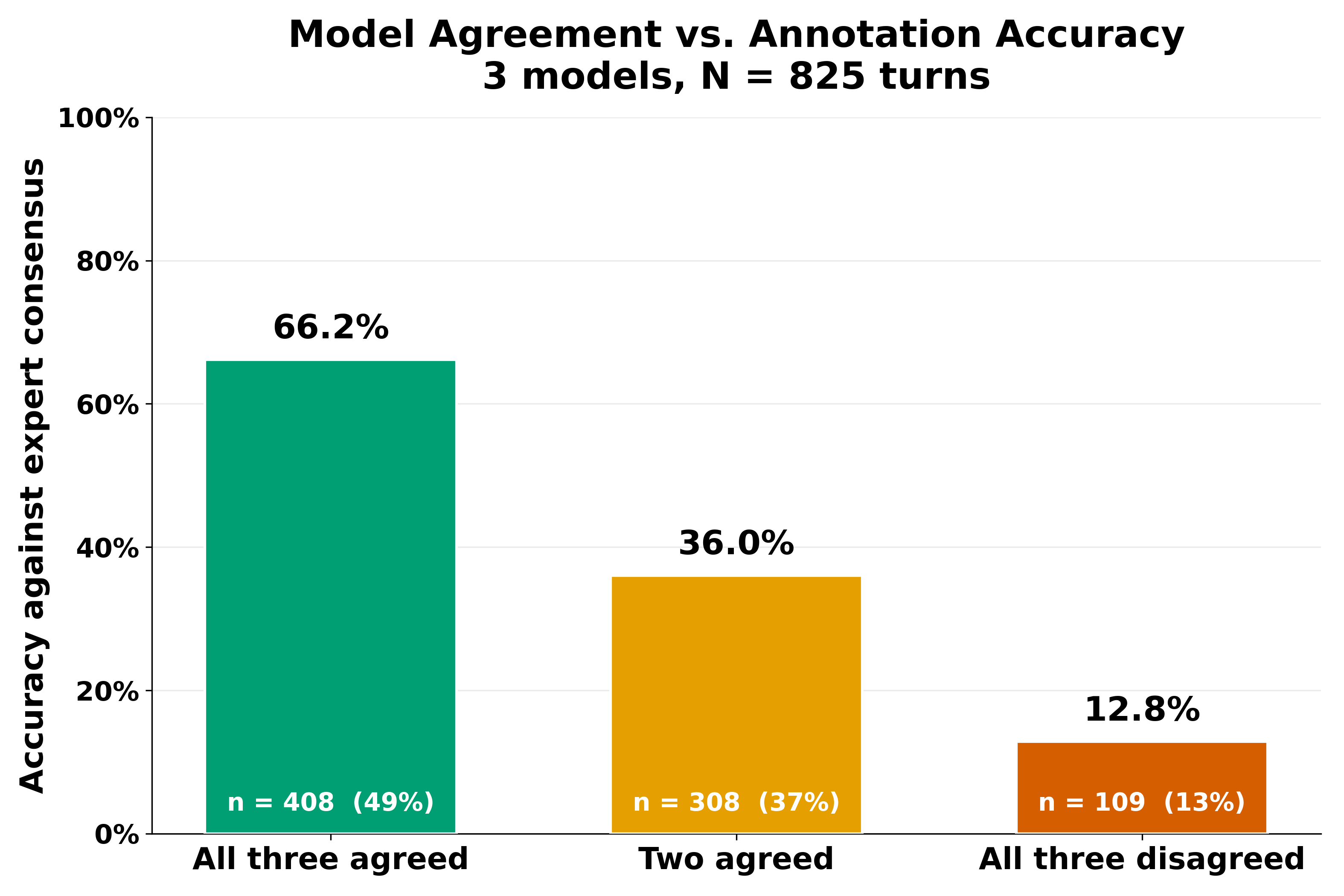}
    \caption{Annotation accuracy by inter-model disagreement on the 825 initial-stage utterances. Predictions from GPT-5.1, Claude Sonnet 4.5, and Gemini 2.5 Pro were aggregated using majority vote and grouped by normalized entropy. Accuracy generally decreases as disagreement increases, indicating that inter-model disagreement provides a useful signal of difficult annotation cases.}
    \label{fig:pre-study-2-initial-data}
    \Description{Bar chart showing annotation accuracy against expert consensus for 825 utterances, grouped by agreement among three models. Accuracy decreases from 66.2\% when all three agree (408 utterances), to 36.0\% when two agree (308 utterances), and 12.8\% when all three disagree (109 utterances).}
\end{figure}

\paragraph{Cross-LLM disagreement is more informative when predictions come from LLMs across different providers.}
Before the main study, we explored how to obtain diverse LLM predictions for measuring disagreement.
We first tested a persona approach that used expert profile information, including their writing and PhD dissertations, to prompt LLMs to mimic individual experts.
However, these expert personas produced relatively little variation in predictions.
In contrast, predictions varied substantially more across LLMs from different providers than across LLMs from the same provider.
This finding motivated our use of six LLMs from six different providers in the main study.
Appendix~\ref{sec:prelim-study} reports the detailed results.

\paragraph{Cross-LLM disagreement effectively identified difficult annotation cases.}
We then tested whether cross-LLM disagreement could identify difficult annotation cases.
We used GPT 5.1, Claude Sonnet 4.5, and Gemini 2.5 Pro to independently annotate the 825 utterances experts labeled during development of the Initial Codebook and aggregated their predictions by majority vote.
With three LLMs, each utterance had one of three agreement patterns: all three agreed, two agreed, or all three disagreed.
Figure~\ref{fig:pre-study-2-initial-data} shows that annotation accuracy decreased as disagreement increased, providing preliminary evidence that cross-LLM disagreement can identify difficult annotation cases.


\section{Discussion}

\subsection{What Roles Can Humans Play in Future Large-Scale Data Annotation?}
Since ChatGPT became publicly available in 2022, a central question in large-scale data annotation has been what role humans should play when LLMs can perform many annotation tasks well.
Early work compared LLMs with individual online crowd workers and often found LLMs more effective~\cite{doi:10.1073/pnas.2305016120,tornberg2023chatgpt}.
Later work compared LLMs with more complete crowdsourcing pipelines, including worker monitoring, label cleaning, and label aggregation, and still found strong LLM performance~\cite{10.1145/3613904.3642834}.
As LLMs increasingly handle simpler annotation tasks that once required large numbers of crowd workers or volunteers, the focus is shifting toward two questions: how to adapt individual LLMs or teams of LLMs for more complex, expert-level annotation, and how to use a limited amount of high-quality expert input to guide them.

Our work contributes to this shift by showing that a small amount of expert input can help LLMs revise a codebook and ultimately outperform a codebook developed manually by experts.
However, our results also suggest that simply asking experts to directly edit or verify codebook revisions may not be the best use of their time.
Combined with prior findings on iterative human editing of prompts for data annotation~\cite{he2025promptingdark}, our results suggest a different role for humans: rather than spreading expert effort across large amounts of annotation or directly editing AI instructions, we can allocate limited expert effort to difficult cases and elicit feedback in forms that LLMs can use effectively.
Thus, as large-scale annotation becomes increasingly automated, not only \emph{where} expert effort is allocated but also \emph{how} expert knowledge is elicited and provided to LLMs becomes increasingly important.

\subsection{The Evolving Role of Codebooks in LLM Data Annotation}

As humans' role in large-scale data annotation evolves, the role of the codebook is also changing. 
For decades, codebooks were written primarily for human annotators, including scholars, students, crowd workers, and volunteers. 
They translated expert knowledge and consensus into definitions, examples, and decision rules that people could follow. 
Today, codebooks are increasingly written for LLMs to read and apply. 
This shift raises a basic question: which properties of traditional codebooks remain useful when the annotator is an LLM? 
Several desirable properties of codebooks that have been mentioned in Section~\ref{sec:related-work}. 
These benefits remain important even when LLMs perform the annotation. 
However, a codebook written for humans may not be the best codebook for LLMs. Our results echo this: codebooks revised by LLMs using targeted expert input produced better LLM annotation performance than a codebook developed through months of expert revision. In the LLM era, a codebook therefore serves not only as a representation of human expert consensus, but also as a way to communicate that consensus effectively to LLMs.
This creates a new tradeoff.
A codebook optimized for LLMs may be less readable or useful to humans, while a codebook designed for human annotators may not provide the instructions that LLMs use most effectively.
Future work should examine which traditional properties of codebooks should be preserved, which should change for LLMs, and whether the same codebook can effectively serve both humans and LLMs.

\subsection{Does Faster Codebook Revision Cut Short Expert Learning?}

Reducing months of codebook development to hours may come at a cost: experts have less time to learn from the data.
Traditional codebook development requires experts to repeatedly examine cases, discuss disagreements, and refine their understanding.
These activities produce not only a better codebook, but also deeper insight into the data.

This concern does not affect our comparison of CV, QA, and RL.
Our study began only after the experts had completed the original annotation process and the Final Codebook was frozen.
Thus, experts already understood the data and labeling task before participating in our workflows.

The broader concern, however, remains.
Whether faster codebook development is desirable depends on the goal of the project.
If experts already know the domain and the goal is accurate large-scale annotation, reducing codebook development time is beneficial.
Experts can also learn about the data through qualitative analysis, case review, exploratory analysis, or downstream analysis rather than annotation alone.
However, if developing deep, data-grounded expert understanding is itself a goal, reducing experts' engagement with the data may be costly.

RL suggests a possible middle ground.
RL still asks experts to examine and reason about individual cases, but strategically selects difficult cases using cross-LLM disagreement.
Experts may therefore not need to examine large amounts of data to learn from it.
A smaller set of carefully selected difficult cases may expose important ambiguities and boundaries more efficiently.
Whether RL can accelerate expert learning, in addition to codebook revision, is an interesting and important direction for future work.

\section{Limitations}
While our workflow can substantially accelerate codebook revision for large-scale data annotation, our study has several limitations. 
Most importantly, we evaluated the workflow in only one longitudinal annotation project with two experts. 
Annotation projects vary in their data, domains, label sets, and annotation difficulty, so our findings may not generalize to other settings. 
Evaluating more projects is important, although longitudinal projects with sustained access to multiple domain experts are difficult to obtain. 
Second, our label set remained largely fixed throughout codebook development, with most changes occurring within the fields of existing labels. 
Although our workflow could support adding, removing, or restructuring labels, we did not observe such changes. 
The workflow may therefore perform differently when the Initial and Final Codebooks differ substantially in their label structure. 
Third, having only two experts across three workflows required us to strategically assign target labels to reduce learning effects across conditions (Section~\ref{sec:target-label-assign}). 
This design does not completely eliminate potential carryover between workflows. 
Finally, the dataset contained occasional errors, including swapped student and tutor identifiers, incorrect utterance segmentation, and transcription errors. 
We intentionally did not manually correct these errors because they reflect noise present in real-world annotation data, but we did not systematically measure their frequency or their effect on our results.

\section{Conclusion and Future Work}

We presented a human-AI workflow that uses cross-LLM disagreement to strategically allocate expert effort for LLM codebook revision.
Rationale Labeling performed best, producing a codebook that outperformed one developed through months of expert annotation and revision with only a few hours of expert effort.
Our results show that LLMs can help identify where expert effort may be valuable
and use targeted expert input to improve large scale annotation.
Future work should evaluate this approach across more domains, codebooks, experts, and data modalities, and examine how concentrating expert effort on difficult cases affects expert learning.

\begin{acks}
We thank Dr. Jennifer John and Dr. Tamisha Thompson for their expert annotations and valuable contributions to this work. This research is based upon initial work completed under National Science Foundation Grant No. 2321499, and support from the Gates Foundation and the Chan Zuckerberg Initiative. Any opinions, findings, and conclusions or recommendations expressed in this material are those of the authors and do not necessarily reflect the views of the funders.
\end{acks}

\bibliographystyle{Style/ACM-Reference-Format}
\bibliography{Bibtex/main}

\appendix

\section{Preliminary Study}\label{sec:prelim-study}

Before designing our human-AI revision workflows, we conducted two preliminary studies to understand what signals could support codebook revision when expert labels are unavailable. 
We progressively asked whether LLMs could 
(1) recover expert-consensus judgments through persona prompting and simulated deliberation, 
and (2) improve the codebook through automatic prompt optimization without expert labels. 
These studies informed our use of diverse models and our decision to elicit targeted expert input in the main study.

\paragraph{Data and evaluation.}
These analyses draw on the longitudinal annotation project described in the main text.
The \emph{Early Expert Annotation Set} contains 825 tutor utterances from 14 sessions, with the experts' independent annotations and subsequent consensus judgments.
Study~1 used this set to evaluate simulated annotation and deliberation.
Study 2 used its utterances to derive revision evidence and evaluate the resulting codebooks on 586 utterances annotated during an intermediate stage of the project.
We refer to the latter as the \emph{preliminary evaluation set} and to the expert-developed snapshot used as an additional reference as the \emph{Intermediate Codebook}. 
The Intermediate Codebook is distinct from the \emph{Final Codebook} used as a reference in the main study. 
The preliminary evaluation should also be distinguished from the main study's evaluation on 4,032 utterances.

We evaluate predictions against the experts' recorded consensus labels, using the first recorded label when a consensus annotation contains multiple labels, consistent with the main study's evaluation convention.
Consensus labels serve as an evaluation reference and they are not newly assigned using a later codebook.
Except for the expert-initialized debate condition in Study~1 and the expert-label-informed revision reference in Study~2, expert annotations are withheld from the simulated annotators and revision procedures.
Thus, the absence of expert labels is a constraint on particular procedures, rather than a description of every preliminary condition.


\subsection{Preliminary Study 1: Simulating Expert Annotation and Deliberation}
\label{sec:prelim-study-1}
We first examined whether persona prompting and simulated deliberation could recover the consensus judgments reached by the two experts. Because the early annotation data preserve both the experts' independent labels and their subsequent consensus, they support a direct comparison between simulated debate outcomes and the experts' recorded decisions.


\paragraph{Deliberation initialized with expert labels.}
We selected the 336 utterances on which the experts initially disagreed. For each utterance, we instantiated two GPT-4o agents, initialized each with the label originally selected by the corresponding expert, and asked them to deliberate for up to 10 rounds to reach a single decision. We compared a persona-conditioned condition, in which each agent additionally received a persona extracted from the corresponding expert's Ph.D. thesis, with a condition without persona prompting. Debates matched expert consensus on 47.6\% of cases with personas and 52.1\% without personas. Debates that remained unresolved after 10 rounds---36 with personas and 38 without---were counted as incorrect. Even with access to the experts' original labels, simulated deliberation therefore recovered approximately half of their consensus judgments; persona conditioning did not improve accuracy in this comparison.

\paragraph{Annotation and deliberation without expert labels.}
We next asked whether the same approach could recover expert consensus without access to the experts' independent labels.
GPT-4o separately annotated the 825 early utterances under each expert-derived persona, achieving 40.2\% and 41.3\% accuracy against consensus. 
The two persona-conditioned annotators disagreed on 110 utterances. Applying the same debate procedure resolved 100 of these cases and achieved 21.8\% accuracy across the 110-utterance disagreement subset.
Combining the initially agreed predictions with the debate outcomes yielded 41.3\% accuracy across all 825 utterances. Thus, the combined procedure did not exceed the more accurate individual persona-conditioned annotator. This comparison evaluates agreement with expert consensus, rather than fidelity to either expert's individual judgments.

\begin{figure*}
    \centering
    \includegraphics[width=0.9\linewidth]{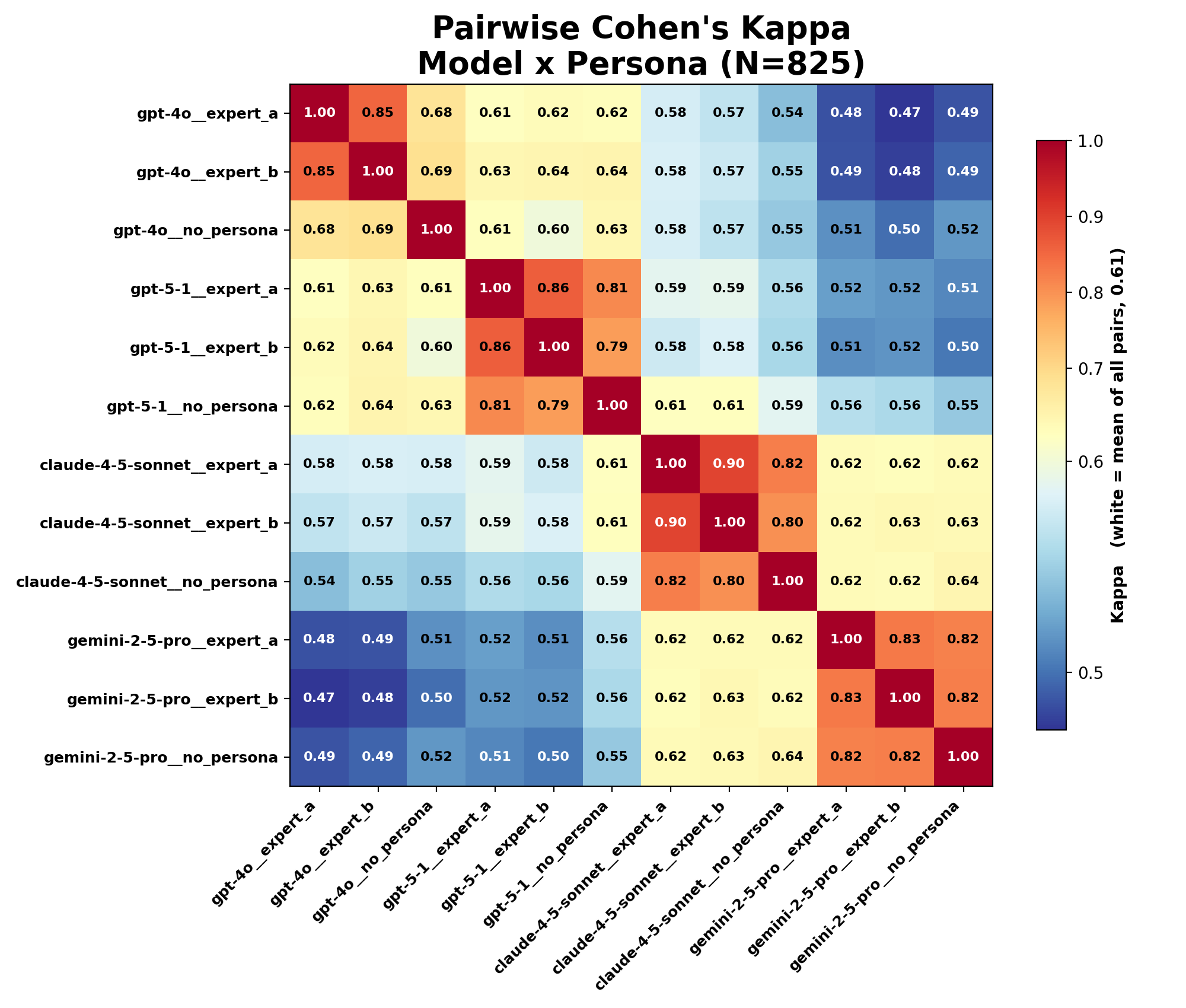}
    \caption{Pairwise Cohen's $\kappa$ across model-persona annotation configurations. Agreement clusters more strongly by the underlying model/provider than by expert persona: configurations using the same model remain relatively similar across persona conditions, whereas applying the same persona to different models does not produce comparable agreement.}
    \Description{Heatmap of pairwise Cohen's kappa among 12 annotation configurations on 825 utterances: four models, each using Expert A’s persona, Expert B’s persona, or no persona. Warmer colors indicate higher agreement. Agreement is higher between configurations using the same model (0.68-0.90) than between different models (0.47-0.64), showing stronger clustering by model than by persona.}
    \label{ref:prel-study-pair-kappa}
\end{figure*}

\paragraph{Model identity and persona conditioning.}
Finally, GPT-4o, GPT-5.1, Claude Sonnet 4.5, and Gemini 2.5 Pro each annotated the same 825 utterances under three conditions: Expert~A's persona, Expert~B's persona, and no persona. Pairwise Cohen's $\kappa$ values across these 12 configurations showed that predictions were more similar across persona conditions within the same base model than across different models using the same persona (Figure~\ref{ref:prel-study-pair-kappa}).
Within-model agreement across persona conditions ranged from approximately $\kappa=0.68$ to $0.90$, whereas agreement between different models using the same expert persona ranged from approximately $\kappa=0.48$ to $0.64$.

These findings suggest that, in the tested configurations, changing the underlying model yielded more variation in annotation behavior than changing the expert persona.
Together with the deliberation results, they motivated our use of multiple models from different providers to obtain diverse predictions in the main study. They do not establish that model identity or provider independently causes a particular annotation pattern.

\subsection{Preliminary Study 2: Automatic Codebook Optimization Without Expert Labels}
\label{sec:prelim-study-3}

We then examined whether established automatic prompt optimization approaches could improve the Initial Codebook without access to expert-consensus labels. 
We adapted SPO~\cite{xiang2025self} and APO~\cite{pryzant2023automatic} to codebook revision under this constraint. 
Randomly sampled high-agreement utterances, paired with model-derived labels, provided pseudo-reference examples. 
Configurations incorporating disagreement additionally received high-entropy utterances as candidate failures.
Agreement-derived labels were treated as model-generated supervision, rather than as verified expert judgments.

On the 586-utterance preliminary evaluation set, the SPO-revised and APO-revised codebooks achieved 48.7\% and 51.2\% accuracy, respectively. 
Neither exceeded disagreement-based revision (51.5\%), and all three were below the Initial Codebook (55.0\%). These results concern the tested adaptations and configurations; they do not establish a general limit on automatic prompt optimization. 
They indicate that substituting model-derived supervision and changing the optimization procedure did not yield an improvement in this setting.

Across the two preliminary studies, cross-model disagreement provided a useful signal for identifying potentially difficult cases. 
Simulated deliberation recovered only a limited proportion of expert-consensus judgments, and the tested revision procedures without expert labels did not improve on the Initial Codebook. 
This motivated the main study's comparison of \emph{Codebook Verifying}, \emph{Question Answering}, and \emph{Rationale Labeling}: three ways to direct limited expert effort toward disagreement and translate that input into codebook revisions.

\section{Label-Level Disagreement Ranking and Expert Assignment}
\label{app:disagreement-entropy-ranking-assignment}

\begin{table*}[t]
\centering

\scriptsize
\resizebox{\textwidth}{!}{
\begin{tabular}{lrrrcccc}
\toprule
\textbf{Label} &
\textbf{$n_{\mathrm{votes}}$} &
\textbf{$n_{\mathrm{inst}}$} &
\textbf{Weighted Entropy} &
\textbf{Pilot} &
\textbf{CV} &
\textbf{QA} &
\textbf{RL} \\
\midrule

\textsc{Revoicing}
& 198 & 140 & 0.6376 & Expert A & Expert B & -- & -- \\

\textsc{Prompting\_Related\_Concepts/Knowledge}
& 371 & 257 & 0.6237 & Expert B & Expert A & -- & -- \\

\textsc{Prompting\_Related\_Representation}
& 441 & 356 & 0.6190 & Expert A & Expert B & -- & -- \\

\textsc{Restating}
& 281 & 182 & 0.6049 & Expert B & Expert A & -- & -- \\

\textsc{Giving\_Hint}
& 846 & 489 & 0.5011 & -- & -- & Expert B & Expert A \\

\textsc{Feedback\_Neutral}
& 946 & 485 & 0.4905 & -- & -- & Expert A & Expert B \\

\textsc{Prompting\_Correction}
& 575 & 277 & 0.4804 & -- & -- & Expert B & Expert A \\

\textsc{Praising\_Processes}
& 212 & 107 & 0.4711 & -- & -- & Expert A & Expert B \\

\textsc{None}
& 2672 & 1158 & 0.4331 & -- & -- & -- & -- \\

\textsc{Asking\_To\_Clarify\_Context}
& 1457 & 678 & 0.4214 & -- & -- & -- & -- \\

\textsc{Giving\_Answer}
& 430 & 178 & 0.4104 & -- & -- & -- & -- \\

\textsc{Encouraging}
& 857 & 385 & 0.3994 & Expert A & -- & Expert A & Expert B \\

\textsc{Giving\_Example}
& 338 & 132 & 0.3901 & -- & -- & Expert B & Expert A \\

\textsc{Strategizing}
& 1350 & 573 & 0.3794 & -- & Expert A & -- & -- \\

\textsc{Technical\_Issues}
& 977 & 376 & 0.3773 & -- & Expert A & -- & -- \\

\textsc{Probing\_Understand}
& 2125 & 1043 & 0.3759 & -- & Expert B & -- & -- \\

\textsc{Prompting\_Self\_Explanation}
& 1489 & 576 & 0.3547 & -- & Expert B & -- & -- \\

\textsc{Asking\_Feeling}
& 627 & 283 & 0.3490 & -- & -- & Expert A & Expert B \\

\textsc{Managing}
& 1796 & 621 & 0.3181 & Expert B & -- & Expert B & Expert A \\

\textsc{Praising\_Traits}
& 28 & 12 & 0.3146 & Expert A & -- & -- & -- \\

\textsc{Correcting\_Own\_Error}
& 438 & 158 & 0.3083 & -- & -- & -- & -- \\

\textsc{Praising\_Outcomes}
& 568 & 217 & 0.3074 & Expert B & -- & -- & -- \\

\textsc{Feedback\_Incorrect}
& 971 & 317 & 0.3028 & -- & -- & Expert B & Expert A \\

\textsc{Probing\_Prior\_Knowledge}
& 778 & 304 & 0.2962 & -- & -- & Expert A & Expert B \\

\textsc{Explaining\_Procedural}
& 3706 & 1168 & 0.2805 & -- & -- & Expert B & Expert A \\

\textsc{Validating\_Feeling}
& 577 & 198 & 0.2749 & -- & -- & Expert A & Expert B \\

\textsc{Prompting\_Action}
& 5839 & 1675 & 0.2542 & Expert A & Expert B & -- & -- \\

\textsc{Explaining\_Conceptual}
& 3075 & 932 & 0.2437 & Expert B & Expert A & -- & -- \\

\textsc{Rapport\_Building}
& 2148 & 591 & 0.2071 & Expert A & Expert B & -- & -- \\

\textsc{Feedback\_Correct}
& 3449 & 933 & 0.1861 & Expert B & Expert A & -- & -- \\

\bottomrule
\end{tabular}
}
\caption{Label-level disagreement statistics and expert assignments for the revision corpus, ranked by vote-weighted entropy. 
$n_{\mathrm{votes}}$ is the total number of votes assigned to the label across the six-model panel, and 
$n_{\mathrm{inst}}$ is the number of utterances for which at least one model predicted the label. 
Pilot, CV, QA, and RL indicate the expert assigned to each label in the corresponding workflow.}
\Description{Table ranking 30 codebook labels by decreasing vote-weighted entropy, from Revoicing (0.6376) to Feedback Correct (0.1861). Columns show model vote counts, utterance counts, and expert assignments for the pilot, CV, QA, and RL workflows. CV uses 12 labels spanning the ranking; QA and RL share 12 different labels, with expert assignments swapped between workflows. Dashes indicate no assignment.}
\label{tab:label-ranking-assignment}
\end{table*}

Table~\ref{tab:label-ranking-assignment} shows the label-level disagreement statistics and expert assignments used in our study, with labels ranked by vote-weighted entropy.
$n_{\mathrm{votes}}$ is the total number of predictions assigned to each label across the six LLMs, while $n_{\mathrm{inst}}$ is the number of utterances for which at least one LLM predicted that label.
The Pilot, CV, QA, and RL columns show which expert was assigned to each label in each workflow.

\section{Prompts and Experimental Settings}\label{app:prompts-experimental-details}
Table~\ref{tab:revision-conditions} summarizes the codebook conditions evaluated in the main study, including the expert contribution, evidence used for revision, and number of labels eligible for revision. The following subsections provide the prompt templates and implementation details used to construct and evaluate these codebooks.

We provide the exact prompt templates and implementation settings used in the main study.
Placeholders indicate where codebook entries, disagreement evidence, and expert input were inserted.
All revision workflows started from the Initial Codebook.
Disagreement evidence was drawn exclusively from the unlabeled revision corpus of 165 sessions containing 6,595 tutor utterances. 
The held-out evaluation sessions and their expert-consensus labels were not available to the revision pipeline.

\begin{table*}[t]
\centering
\footnotesize
\setlength{\tabcolsep}{5pt}
\renewcommand{\arraystretch}{1.2}
\resizebox{\textwidth}{!}{%
\begin{tabular}{@{}lllll@{}}
\toprule
\textbf{Revision approach} & \textbf{Codebook Condition} & \textbf{Expert contribution} &
\textbf{Revision evidence} & \textbf{Labels eligible for revision} \\
\midrule

\textbf{Expert-only} & Initial codebook & None & None & 0 \\
\midrule

\multirow{5}{*}{\textbf{CV}}
 & Fully LLM-revised & None & Model disagreement & 30 \\
\cmidrule(l){2-5}
 & Expert-validated + 
 & Verify/edit proposed
 & \multirow{2}{*}{Model disagreement}
 & \multirow{2}{*}{12} \\
 & initial & revisions & & \\
\cmidrule(l){2-5}
 & Expert-validated +
 & Verify/edit proposed
 & \multirow{2}{*}{Model disagreement}
 & \multirow{2}{*}{30} \\
 & LLM-revised & revisions & & \\
\midrule

\multirow{6}{*}{\textbf{QA}}
 & \multirow{2}{*}{Model disagreement + QA}
 & \multirow{2}{*}{Clarification answers}
 & Model disagreement +
 & \multirow{2}{*}{12} \\
 & & & QA & \\
\cmidrule(l){2-5}
 & QA-only & Clarification answers
 & QA & 12 \\
\cmidrule(l){2-5}
 & Ambiguity-pattern + QA & Clarification answers
 & Ambiguity pattern + QA & 12 \\
\cmidrule(l){2-5}
 & Ambiguity-pattern + QA +
 & \multirow{2}{*}{Clarification answers}
 & Ambiguity pattern + & \multirow{2}{*}{25} \\
 & Cross-label QA & & within- and cross- label QA & \\
\midrule

\multirow{2}{*}{\textbf{RL}}
 & \multirow{2}{*}{RL-revised}
 & \multirow{2}{*}{Labels + rationales}
 & Expert label +
 & \multirow{2}{*}{24} \\
 & & & reasoning\\
\midrule

\multirow{2}{*}{\textbf{Expert-only}}
 & \multirow{2}{*}{Final codebook}
 & Full expert annotation,
 & \multirow{2}{*}{Expert longitudinal process}
 & \multirow{2}{*}{30} \\
 & & discussion, and revision & & \\
\bottomrule
\end{tabular}}
\caption{Codebook conditions evaluated in the main study. 
``Expert contribution'' describes how experts guide revision through reviewing and editing proposed changes (CV), answering clarification questions (QA), or labeling utterances and providing rationales (RL).
``Revision evidence'' identifies the information used to generate revisions, including disagreement cases and model rationales, diagnosed ambiguity patterns, and expert input, depending on the condition.
All AI-only and human-AI revision conditions start from the Initial Codebook.
GPT-5.6 Sol generates candidate revisions for CV and produces revised entries from expert answers in QA or expert labels and rationales in RL.
``Labels eligible for revision'' denotes the number of codebook labels eligible for change under each condition.
The Initial and Final Codebooks serve as reference snapshots from the independent longitudinal expert development process.
}
\Description{Table summarizing ten codebook conditions by revision approach, expert contribution, revision evidence, and number of labels eligible for revision. CV involves reviewing and editing proposed revisions; QA uses expert clarification answers; RL uses expert labels and rationales. Conditions incorporate different combinations of model disagreement, ambiguity patterns, and expert input. Initial and final expert-developed codebooks serve as reference conditions.}
\label{tab:revision-conditions}
\end{table*}

\subsection{Data and Initial Annotation}
All three workflows started from the Initial Codebook and used the same unlabeled revision corpus of 165 sessions containing 6,595 tutor utterances. 
The held-out evaluation set contained 4,032 expert-consensus tutor utterances from 110 separate sessions. Neither these evaluation sessions nor their expert-consensus labels were available to the revision pipeline. 

We first used six models to independently annotate the revision corpus with the Initial Codebook:
\begin{itemize}
    \item \texttt{openai.gpt-5.1.2025-11-13}
    \item \texttt{google.gemini-2.5-pro}
    \item \texttt{anthropic.claude-4.5-sonnet}
    \item \texttt{qwen.qwen3-coder-480b-a35b}
    \item \texttt{deepseek.v3.2}
    \item \texttt{mistral.mistral-large-3-675b-instruct}.
\end{itemize}

The utterance annotation prompt (Figure~\ref{fig:utterance-annotation-prompt}) supplied the codebook, preceding conversation, and current message. It requested a single label, a short rationale, and an uncertainty rating, returning \texttt{None} for turns matching no tutoring move.
The same prompt template was used for codebook evaluation. 
\begin{promptlisting}{Utterance Annotation}
================ SYSTEM PROMPT ================

You are an expert coding agent annotating tutoring session messages. Your task is to assign the single best tutoring-move label from the codebook to the CURRENT MESSAGE given the preceding conversation context.

{codebook_content}

Rules:
- Only annotate the CURRENT MESSAGE, not any other message in the conversation.
- Use the full conversation as context to understand the current message.
- If the message is from a student or does not match any tutoring move, use "None".

OUTPUT FORMAT
Respond with ONLY valid JSON - no markdown fences, no extra text:
{{
  "current_label": "<exact codebook label name>",
  "reasoning": "<2-4 sentences referencing codebook definitions>",
  "uncertainty_level": "<low/medium/high>",
  "flagged_for_review": false
}}

================= USER PROMPT =================
CONVERSATION HISTORY:
{conversation_history}

CURRENT MESSAGE TO ANNOTATE:
{role}({timestamp}): {message}

\end{promptlisting}
\captionof{figure}{User and system prompt templates for utterance annotation, shared across all conditions. Only the codebook inserted at \texttt{\{codebook\_content\}} varies across conditions. The prompt supplies the preceding conversation as context and identifies the target utterance by speaker role and timestamp. The model returns a single label, its reasoning, and an uncertainty rating, assigning \texttt{None} to utterances that match no tutoring move.} 
\label{fig:utterance-annotation-prompt}
\Description{Text panel showing system and user prompt templates for utterance annotation. The model receives a codebook, preceding conversation, and target message, then returns a single label, reasoning, uncertainty level, and review flag in JSON format. Utterances matching no tutoring move receive ``None.'' Only the supplied codebook varies across conditions.}


\paragraph{Cross-Model Disagreement Calculation, Assignment, and Selection}\label{app:sec-utt-selection}
Using the six models' predictions, we calculated utterance-level disagreement $D(u)$ as normalized vote entropy and class-level disagreement $D(c)$ as the vote-weighted mean of utterance-level disagreement (Section~\ref{sec:disagreement-method}). 
Class-level scores determined the ranking used to assign target labels to experts, whereas utterance-level scores guided evidence selection. 

\textbf{Target Label Assignment.} 
Based on the class-level disagreement ranking (Table~\ref{tab:label-ranking-assignment}), we selected the top four, middle four, and bottom four labels as CV target labels.
QA and RL shared a separate set of 12 target labels adjacent to the CV selections: four immediately below the top four, two immediately above and two immediately below the middle four, and four immediately above the bottom four. 
Each workflow's target labels were divided equally between the two experts, with assignments swapped between QA and RL so that no expert was assigned the same target label more than once across the three workflows.

\textbf{Utterance Selection for CV and QA.} 
We constructed a candidate pool of utterances across all 30 labels. Each selected utterance had at least two distinct labels predicted for it, with at least one of the six models predicting the specific label in question.
We ranked eligible utterances by normalized vote entropy $D(u)$ in descending order and selected up to 30 utterances per label.
These utterances, together with the six models' predictions and rationales, supported CV revision generation (Section~\ref{app:sec-cv}) and QA disagreement-QA revision (Section~\ref{app:sec-qa}). 
RL used a separate selection procedure, described below.

\textbf{Utterance Selection for RL.}
For each RL target label, we constructed a candidate pool containing utterances for which at least one model predicted that label.
We selected ten utterances per target label using a greedy procedure: at each step, we selected the utterance introducing the most competing labels not yet represented in the selected set, breaking ties by prioritizing higher $D(u)$.
Target labels were processed in ascending order of eligible pool size to preserve available examples for labels with fewer candidates. Each utterance was assigned to only one RL target label across the study.

\subsection{Codebook Verifying}\label{app:sec-cv}
\paragraph{Generating Candidate Revisions}
We used \texttt{gpt-5.6-sol} to generate candidate revisions for all 30 labels (Figure~\ref{fig:cv-revision-system-prompt} and \ref{fig:cv-revision-user-prompt}).
For each label, the revision prompt supplied its Initial Codebook entry and the selected disagreement utterances, including the predictions and rationales from all six annotation models. The model proposed revisions to five fields: Explanation, Examples, Near Hits, Near Misses, and Non Examples. 
No expert input was provided at this stage. 

\begin{promptlisting}{CV: codebook revision (1/2)}
================ SYSTEM PROMPT ================
You are an expert in educational research, learning science, and tutoring codebook design.

Your job is to produce a *conceptual* revision of a tutoring-move codebook entry using model annotation evidence --- no expert ground truth is available.

The evidence comes in ONE form only:
  * HIGH-DISAGREEMENT EXAMPLES: items where models split across labels, with each model's reasoning explaining why it chose what it chose. These reveal exactly where the codebook definition is under-specified or ambiguous.
No agreement (confirming) examples are provided.

Guidelines:
  - Explanation: Articulate the underlying educational or cognitive mechanism ---
    what this move IS and WHY it works, not just what it looks like on the surface.
    Derive precise boundary conditions directly from the disagreement patterns.
  - Examples: This field should be MOSTLY example utterances, with minimal
    explanation. Keep/refine the ORIGINAL entry's examples, and add verbatim
    utterances from the HIGH-DISAGREEMENT EXAMPLES only where this label clearly
    fits; it is fine to have only a few. Do NOT explain each one; at most add a
    2-4 word tag in parentheses, and only when genuinely helpful. Prioritize
    concrete examples over describing them.
  - Near-Hit (close but not quite): Like Examples, this should be MOSTLY
    utterances with minimal explanation. Select as many utterances as the
    HIGH-DISAGREEMENT EXAMPLES support where this label won (majority); fewer is
    fine when little data is available. Do NOT write a full sentence per item; at
    most add a short tag in parentheses naming the competing label, e.g.
    "(vs OTHER_LABEL)".
  - Near-Miss (yes, but almost not): Like Examples, MOSTLY utterances with
    minimal explanation. Write plausible utterances that satisfy most criteria
    but fail on one specific conceptual dimension. At most a short tag in
    parentheses naming the missing dimension --- no full sentences.
  - Non-Examples: Like Examples, MOSTLY utterances with minimal explanation.
    Write utterances that clearly do NOT qualify. At most a short tag in
    parentheses naming the label they actually belong to --- no principled-reason
    sentences.

Output ONLY valid JSON --- no markdown, no extra text.
\end{promptlisting}

\captionof{figure}{System prompt template for generating LLM-proposed codebook revisions in the CV workflow. The model revises an original codebook entry using cross-model disagreement and model rationales, without expert ground-truth annotations.}
\Description{System prompt for generating CV codebook revisions from cross-model disagreement and model rationales, without expert ground-truth annotations.}
\label{fig:cv-revision-system-prompt}

\begin{promptlisting}{CV: codebook revision (2/2)}
================= USER PROMPT =================

You are revising the codebook entry for tutoring-move label **{label}**.

No expert ground truth is available. You have model annotation evidence instead ---
ONLY disagreement cases (no agreement examples are provided).

=== ORIGINAL CODEBOOK ENTRY ===
LABEL: {label}
CATEGORY: {category}

## Current definition
{target_label_definition}

=== HIGH-DISAGREEMENT EXAMPLES ===
{disagreement_block}

Produce a single revised JSON object with the keys below. For the three definition-role keys (Definition / Expanded Definition / Explanation), include ONLY the ones that appear in the ORIGINAL CODEBOOK ENTRY above and omit the others; all remaining keys are always required.

{
  "Explanation": "<Conceptually grounded explanation. State the educational or cognitive mechanism this move serves. Include precise boundary conditions derived from the disagreement patterns above --- which labels does it most frequently get confused with, and what is the conceptual distinguishing factor?>",

  "Examples": "<This field should be MOSTLY example utterances, with as little explanation as possible. Keep/refine the ORIGINAL entry's examples, and add utterances verbatim from the HIGH-DISAGREEMENT EXAMPLES above only where {label} clearly fits; fewer is fine when little data is available. Do NOT write a sentence explaining each one --- at most append a short 2-4 word tag in parentheses, and only when genuinely useful. Separate entries with ' | '.>",

  "Near-Hit (close but not quite)": "<MOSTLY utterances, minimal explanation. Select as many utterances verbatim from the HIGH-DISAGREEMENT EXAMPLES where {label} is the majority label but models split as the evidence supports; fewer is fine when little data is available. Do NOT write a sentence per item --- at most append a short tag in parentheses naming the competing label, e.g. '(vs OTHER_LABEL)'. Separate entries with ' | '.>",

  "Near-Miss (yes, but almost not)": "<MOSTLY utterances, minimal explanation. Write plausible utterances that satisfy most criteria of {label} but fail on one specific conceptual dimension. At most a short tag in parentheses naming the missing dimension --- no full sentences. Separate entries with ' | '.>",

  "Non-Examples": "<MOSTLY utterances, minimal explanation. Write utterances that clearly do NOT qualify as {label}. At most a short tag in parentheses naming the label they actually belong to --- no principled-reason sentences. Separate entries with ' | '.>",

  "conceptual_boundary_summary": "<2-3 sentences articulating the core conceptual boundary between {label} and the 1-2 labels it is most frequently confused with, grounded in the disagreement evidence above.>"
}

Output JSON only.
\end{promptlisting}

\captionof{figure}{User prompt template for generating LLM-proposed codebook revisions in the CV workflow. The template supplies the target label (\texttt{\{label\}}), its original codebook entry (\texttt{\{original\_entry\}}), and high-disagreement examples with model predictions and rationales (\texttt{\{disagreement\_block\}}). It requests a revised entry and a summary of its conceptual boundaries with commonly confused labels.}
\Description{User prompt for CV codebook revision, supplying the original label entry and disagreement evidence and requesting a revised entry with clarified label boundaries.}
\label{fig:cv-revision-user-prompt}

\paragraph{Expert Validation and Codebook Construction}
Experts reviewed the proposed revisions for 12 assigned target labels, six per expert, approving, editing, or rejecting changes field by field.
We constructed three codebook conditions: \textit{Fully LLM-revised}, which used the proposed revisions for all labels without expert validation; \textit{Expert-validated + initial}, which combined expert-validated entries for target labels with initial entries for unreviewed labels; and \textit{Expert-validated + LLM-revised}, which combined expert-validated entries for target labels with LLM-revised entries for unreviewed labels.

\subsection{Question Answering}\label{app:sec-qa}
\paragraph{Extracting Ambiguity Patterns.}
We used \texttt{gpt-5.6-sol} to identify recurring ambiguity patterns from each target label's Initial Codebook entry, competing label definitions, summary statistics of disagreement, and selected disagreement utterances with model predictions and rationales (Figure~\ref{fig:qa-ambiguity-system-prompt} and \ref{fig:qa-ambiguity-user-prompt}).
Competing label definitions were included in descending order of alternative-choice probability, using the smallest set whose cumulative probability reached at least 50\%. 
The prompt asked the model to identify unresolved criteria, boundaries, and scope rather than adjudicate individual utterances. 

\begin{promptlisting}{QA: extracting ambiguity patterns (1/2)}
============================== SYSTEM PROMPT ==============================

You are an expert in educational research, learning science, and the design of qualitative annotation codebooks for tutoring dialogue.

You are diagnosing a codebook, not annotating data. You are shown situations in which several careful, independent readers of the same codebook entry applied it differently --- each convinced they were right. Your task is to explain what about the CODEBOOK, not about the readers, allowed that to happen.

Work by ABSTRACTION. Do not adjudicate individual cases and do not decide which reading was correct. Look across the cases for RECURRING structures: a condition the entry never states, a boundary it never draws, a scope it leaves open, a distinction it implies but never operationalizes.

A good pattern:
  - names a decision the annotator must make but the codebook does not resolve;
  - is stated in general terms, so it applies to situations beyond those shown;
  - is grounded in more than one case;
  - identifies exactly which competing labels are at stake, when relevant.

A bad pattern:
  - restates one example;
  - says only "the definition is vague";
  - blames annotator carelessness;
  - describes a distinction the codebook entry ALREADY draws explicitly.

Be honest when the entry does already resolve something: mark it as such rather than inventing a gap.

Output ONLY valid JSON --- no markdown, no commentary.
\end{promptlisting}
\captionof{figure}{System prompt template for QA \textbf{ambiguity-pattern extraction}, which frames the task as diagnosing the codebook rather than the annotators.}
\Description{System prompt for identifying recurring codebook ambiguities from model disagreement to inform the QA workflow.
}
\label{fig:qa-ambiguity-system-prompt}

\begin{promptlisting}{QA: extracting ambiguity patterns (2/2)}
=============================== USER PROMPT ===============================

Diagnose the codebook entry for the tutoring-move label **{label}**.

=== CURRENT CODEBOOK ENTRY (the text annotators actually read) ===
{target_label_definition}

=== ENTRIES FOR THE LABELS MOST OFTEN APPLIED INSTEAD ===
{competing_labels}

=== WHAT READERS CHOSE INSTEAD OF {label}, AND HOW OFTEN ===
{competing_labels_stats}

=== SITUATIONS WHERE READERS DISAGREED ===
{disagreement_block}

=== YOUR TASK ===
Identify the recurring ambiguities in the CURRENT entry for {label} that could explain these divergent readings. Produce between {min_patterns} and {max_patterns} patterns, ordered by how much of the disagreement they explain.

Return exactly this JSON object:

{
  "label": "{label}",
  "ambiguity_summary": "<3-5 sentences: the overall diagnosis of what makes this entry hard to apply consistently.>",
  "dominant_ambiguity_type": "<label_internal if the criteria for applying {label} at all are unclear | cross_label if the trouble is one specific rival label | mixed>",
  "patterns": [
    {
      "pattern_id": "P1",
      "title": "<short noun phrase naming the ambiguity, max 12 words>",
      "type": "<label_internal | cross_label>",
      "competing_labels": ["<labels at stake; [] for label_internal>"],
      "description": "<The recurring situation, stated abstractly and generally --- a class of cases, never one utterance. 2-4 sentences.>",
      "unresolved_decision": "<The specific decision an annotator must make that the entry leaves open, phrased as a decision: 'whether X counts when Y'.>",
      "why_readers_diverge": "<Which cue each side is keying on, and why the current wording licenses both readings.>",
      "codebook_gap": "<one of: {gap_types}>",
      "current_codebook_coverage": "<absent if the entry says nothing about this | partial if it gestures at it without a decidable rule | explicit if the entry already resolves it and the disagreement is a misreading>",
      "prevalence": "<high | medium | low --- how much of the observed disagreement this pattern accounts for>",
      "evidence_item_keys": ["<item_key of each supporting case above>"]
    }
  ]
}

Output JSON only.
\end{promptlisting}

\captionof{figure}{User prompt template for QA \textbf{ambiguity-pattern extraction}, which supplies the target codebook entry, competing-label entries and frequencies, and examples of cross-model disagreement.}
\Description{
User prompt for identifying recurring codebook ambiguities from model disagreement to inform the QA workflow.
}
\label{fig:qa-ambiguity-user-prompt}

\paragraph{Generating Questions and Collecting Expert Answers.}
A second prompt, also run with \texttt{gpt-5.6-sol}, converted the ambiguity patterns into six self-contained questions per target label, addressing core behavior, necessary conditions, exclusions, boundaries, scope, and edge cases (Figure~\ref{fig:qa-generate-question-system-prompt} and \ref{fig:qa-generate-question-user-prompt}).
Experts answered questions for 12 target labels, six per expert. Each expert answered 36 questions, yielding 72 responses in total. 
Experts did not see the underlying transcripts, model predictions, rationales, or ambiguity analysis. 

\begin{promptlisting}{QA: generating clarification questions (1/2)}
============================== SYSTEM PROMPT ==============================

You are an expert in educational research and the design of qualitative annotation codebooks for tutoring dialogue.

You write clarification questions for a senior domain expert --- the person whose judgment the codebook is meant to encode. You have a diagnosis of where the codebook is ambiguous. Your job is to turn that diagnosis into questions whose answers would let someone rewrite the codebook entry so the ambiguity disappears.

The expert will see ONLY your questions. They will not see any transcript, any prior labeling, or anything about how the ambiguity was found. So:

  - Every question must stand entirely on its own. Never refer to an example, a case, a transcript, or anything "above" --- there is nothing above.
  - Ask for a REUSABLE DECISION RULE, not the right answer for one utterance. "How should one decide whether X counts when Y?" --- not "is this X?"
  - Write in plain domain language: tutoring, students, moves, codes, categories. Never mention automated annotation, statistics, uncertainty scores, or how the ambiguity was detected. The expert should read a thoughtful colleague's question about the codebook.
  - Keep every question CENTRED ON THE TARGET CATEGORY. You may name a neighboring category when that is the clearest way to ask about a boundary, but the name must sharpen the question, not carry it: always also DESCRIBE the confusable behavior concretely --- what the tutor is doing and what the student gets from it --- so the question stands on its own even for a reader who knows the neighboring category only loosely. "How should a turn that states the underlying reason a rule holds, rather than walking through the steps of applying it, be separated from EXPLAINING_PROCEDURAL?" --- not "how does this differ from EXPLAINING_PROCEDURAL?".
  - Do not ask what the codebook already answers explicitly. If the entry already states a rule, do not re-ask for it.
  - Scenario questions must be SYNTHESISED: invent a clean, typical situation that isolates the ambiguity. Never reproduce a real utterance.
  - Each question must be answerable by an expert in a few sentences, and its answer must translate into concrete codebook text --- a definition, an inclusion or exclusion criterion, a boundary rule, an edge case, or an example.

Prefer a few sharp questions that surface the expert's latent decision rules over many shallow ones. Mix broad conceptual questions with concrete detailed ones.

Output ONLY valid JSON --- no markdown, no commentary.
\end{promptlisting}
\captionof{figure}{System prompt template for QA \textbf{clarification-question generation}, which turns the diagnosed ambiguities into questions for a senior domain expert.}
\Description{System prompt for turning diagnosed codebook ambiguities into clarification questions that elicit reusable decision rules from experts in the QA workflow.}
\label{fig:qa-generate-question-system-prompt}

\begin{promptlisting}{QA: generating clarification questions (2/2)}
=============================== USER PROMPT ===============================
Write clarification questions about the codebook entry for **{label}**.

=== THE CURRENT CODEBOOK ENTRY THE EXPERT WROTE ===
{target_label_definition}

=== NEIGHBORING CATEGORIES READERS REACHED FOR INSTEAD ===
These are the categories most often chosen instead of {label} on contested
turns. Use their definitions to draw precise boundaries; you may name them in
questions where that sharpens the contrast.

{competing_labels}

=== DIAGNOSED AMBIGUITIES IN THIS ENTRY ===
{ambiguity_summary}

=== YOUR TASK ===
Produce EXACTLY {n_questions} questions --- one for each aspect of the definition listed below, in this order, and no more than one per aspect. Each is the only question that aspect will get, so make it the best one you can rather than the first that occurs to you.

  core_behaviour     what the move fundamentally IS --- what the tutor is doing and what it does for the student, beneath any particular wording
  necessary_condition  what must be present for the category to apply at all, as opposed to what is merely typical of it
  exclusion          what closely resembles the category but must NOT be coded as it, and on what grounds
  boundary           where this category stops and a neighboring one begins, for the confusion the diagnosis found most often
  scope              how to treat a turn that does several things at once, or where the behavior appears only briefly inside a longer turn
  edge_case          the awkward situations a careful annotator will meet that the entry does not settle --- degenerate, borderline, or rare-but-recurring

Do not spend two questions on the same aspect, however much material there is for it: an entry whose boundary is asked about four times and whose core is never asked about leaves the expert unable to rebuild the definition. If an aspect looks well settled by the current entry, still ask about it --- aim the question at whatever is least settled within that aspect.

Ground every question in the diagnosis: cite the pattern it comes from in `pattern_id`, and between them address the ambiguities identified above rather than inventing new ones.

Return exactly this JSON object:

{
  "label": "{label}",
  "questions": [
    {
      "question_id": "Q1",
      "pattern_id": "<the pattern this addresses>",
      "type": "<{types}>",
      "definition_aspect": "<which aspect of the definition this probes: {aspects}>",
      "granularity": "<conceptual | detailed>",
      "question": "<The question as the expert will read it. Self-contained, plain domain language, seeking a reusable rule. Centered on {label}; other categories may be named where it sharpens a boundary, but always alongside a concrete description of the behavior --- say what the tutor does and what the student gets from it. For a scenario question, give the invented situation first in 1-2 sentences, then the rule question.>",
      "why_it_matters": "<One sentence, for the research team only: which ambiguity this resolves. Not shown to the expert.>",
      "expected_revision_target": "<what the answer would let us write: one of {targets}>",
      "related_labels": ["<other codebook labels the answer would also affect; [] if none>"],
      "answerable_from_current_codebook": <true if the entry above already answers this and it should therefore be dropped, else false>
    }
  ]
}

Output JSON only.
\end{promptlisting}

\captionof{figure}{User prompt template for QA \textbf{clarification-question generation}, supplying the entry, the neighboring categories readers chose instead, and the diagnosed ambiguity patterns.}
\Description{User prompt for turning diagnosed codebook ambiguities into clarification questions that elicit reusable decision rules from experts in the QA workflow.}
\label{fig:qa-generate-question-user-prompt}

\paragraph{Revising the Codebook from Expert Questions and Answers.}
We constructed four codebook conditions: 
\textit{Disagreement + QA}, which revised the 12 target labels using selected disagreement utterances, model reasoning, and expert questions and answers, with expert answers taking precedence over model reasoning (Figure~\ref{fig:qa-revision-disagree-qa});
\textit{QA only}, which revised the 12 target labels using expert questions and answers without additional disagreement evidence (Figure~\ref{fig:qa-revision-qa-only});
\textit{Ambiguity Pattern + QA}, which revised the 12 target labels using ambiguity patterns and associated disagreement context alongside expert questions and answers, treating expert answers as the only source of revision rules (Figure~\ref{fig:qa-revision-ambiguity-qa});
and \textit{Ambiguity Pattern + Within and Cross Label QA}, which additionally incorporated relevant expert questions and answers about other labels, allowing revision of both target labels and other labels explicitly involved in expert responses (Figure~\ref{fig:qa-revision-cross-qa-system} and \ref{fig:qa-revision-cross-qa-user}).
The fourth condition gave direct answers precedence over cross-label answers and resulted in 25 revised entries. In all conditions, labels not eligible for revision retained their Initial Codebook entries.

\begin{promptlisting}{QA codebook revision: disagreement + QA}
============================== SYSTEM PROMPT ==============================
You are an expert in educational research, learning science, and tutoring codebook design.

Your job is to produce a *conceptual* revision of a tutoring-move codebook entry using two kinds of evidence: model annotation evidence, and an expert annotator's written answers to clarification questions about this label.

The evidence comes in TWO forms, and they do NOT rank equally:
  * EXPERT ANSWERS: a human expert's written rulings on questions asked about this label. These are authoritative. Where an answer settles something, that settlement is the rule, and where an answer conflicts with a model's reasoning the expert is right and the model is wrong.
  * HIGH-DISAGREEMENT EXAMPLES: items where several independent readers of this codebook entry split across labels, each with the reasoning behind their choice. They show WHERE the definition is under-specified, and they supply the real utterances you must quote. They do not settle anything by themselves, and the readers are deliberately unnamed: what matters is that the entry permitted the split, not who made which call. No agreement (confirming) examples are provided.

Where the expert did not address something the disagreement evidence raises, you may still sharpen the wording, but do not invent a rule the expert would have had to decide.

Guidelines:
  - Explanation: Articulate the underlying educational or cognitive mechanism ---
    what this move IS and WHY it works, not just what it looks like on the surface.
    Derive precise boundary conditions directly from the disagreement patterns.
  - Examples: This field should be MOSTLY example utterances, with minimal
    explanation. Keep/refine the ORIGINAL entry's examples, and add verbatim
    utterances from the HIGH-DISAGREEMENT EXAMPLES only where this label clearly
    fits; it is fine to have only a few. Do NOT explain each one; at most add a
    2-4 word tag in parentheses, and only when genuinely helpful. Prioritise
    concrete examples over describing them.
  - Near-Hit (close but not quite): Like Examples, this should be MOSTLY
    utterances with minimal explanation. Select as many utterances as the
    HIGH-DISAGREEMENT EXAMPLES support where this label won (majority); fewer is
    fine when little data is available. Do NOT write a full sentence per item; at
    most add a short tag in parentheses naming the competing label, e.g.
    "(vs OTHER_LABEL)".
  - Near-Miss (yes, but almost not): Like Examples, MOSTLY utterances with
    minimal explanation. Write plausible utterances that satisfy most criteria
    but fail on one specific conceptual dimension. At most a short tag in
    parentheses naming the missing dimension --- no full sentences.
  - Non-Examples: Like Examples, MOSTLY utterances with minimal explanation.
    Write utterances that clearly do NOT qualify. At most a short tag in
    parentheses naming the label they actually belong to --- no principled-reason
    sentences.

Output ONLY valid JSON --- no markdown, no extra text.
=============================== USER PROMPT ===============================
You are revising the codebook entry for tutoring-move label **{label}**.

No expert ground truth is available. 

=== ORIGINAL CODEBOOK ENTRY ===
LABEL: {label}
CATEGORY: {category}

## Current definition
{target_label_definition}

## EXPERT ANSWERS
{expert_questions_answers}

## HIGH-DISAGREEMENT EXAMPLES
{high_disagreement_examples}

Output JSON only.
\end{promptlisting}
\captionof{figure}{System and user prompt templates for the \textbf{Disagreement + QA} revision condition. The prompt keeps the same CV's disagreement-driven shape and adds the expert's answers as authoritative evidence.}
\Description{System and user prompts for revising codebook entries using model disagreement and expert clarification answers, with expert answers taking precedence over model reasoning.}
\label{fig:qa-revision-disagree-qa}

\begin{promptlisting}{QA codebook revision: QA only}
============================== SYSTEM PROMPT ==============================

You are a methodologist maintaining a qualitative-coding codebook. An expert annotator has answered targeted questions about ONE label whose definition was found to be ambiguous. Rewrite that label's entry so the ambiguities the questions raised are resolved.

Rules:
- Every change must be grounded in the expert's answers. Never invent a rule, boundary, criterion or example the expert did not state or clearly imply.
- If an answer is vague or non-committal, leave the relevant text as it is rather than guessing at what was meant.
- Each question notes the field it was expected to affect. Treat that as a hint, not an instruction: put what the expert said where it actually belongs.
- Keep every field that already exists, and return unchanged fields verbatim.
- Keep the original voice, format and level of detail. This is a working codebook, not an essay: prefer a sharper sentence over a longer one.
- Examples stay short, quoted utterances in the style already used.
- Do not rename the label and do not change its category.

Reply with a single JSON object and nothing else:
{"revised": {"<field>": "<full text of the field after revision>", ...},
 "changed_fields": ["<field>", ...],
 "rationale": "<2-4 sentences: what you changed and which answers drove it>"}
The "revised" object must contain exactly the field names you were given.

=============================== USER PROMPT ===============================
You are revising the codebook entry for tutoring-move label **{label}**.

No expert ground truth is available. 

=== ORIGINAL CODEBOOK ENTRY ===
LABEL: {label}
CATEGORY: {category}

## Current definition
{target_label_definition}

## Expert answers
{expert_questions_answers}

## Fields to return
["Explanation", "Examples", "Near-Miss (yes, but almost not)", "Near-Hit (close but not quite)", "Non-Examples"]

Revise the entry and reply with the JSON object described in your instructions.
\end{promptlisting}
\captionof{figure}{System and user prompt templates for the \textbf{QA-only} revision condition. The model receives the current definition and the expert's answers. Every change is grounded in what the expert stated or clearly implied.}
\Description{System and user prompts for revising codebook entries using expert clarification answers
}
\label{fig:qa-revision-qa-only}

\begin{promptlisting}{QA codebook revision: Ambiguity-pattern + QA}
============================== SYSTEM PROMPT ==============================
You are a methodologist maintaining a qualitative-coding codebook. An expert annotator has answered targeted questions about ONE label whose definition was found to be ambiguous. You are given the label's current entry, a diagnosis of the ambiguities found in it, the categories readers chose instead, real turns readers disagreed about, and the expert's answers. Rewrite the entry so the ambiguities are resolved.

These sources do NOT carry equal authority.

- THE EXPERT'S ANSWERS are the only source of RULES. Every criterion, boundary, exclusion or priority you write must come from what the expert said or clearly implied. If they did not settle something, leave it unsettled.
- THE DIAGNOSIS says what each question was asking and why readers split. Use it to interpret a terse answer. It is analysis, not evidence, and never a source of rules.
- THE CONTESTED TURNS are real utterances from the corpus, and they are the SAME turns the questions were written from. An answer that says "a bare request can count" is a ruling about the kind of turn shown here, so read the answers against these utterances rather than in the abstract. They are also the source of example TEXT: quote them verbatim, keeping their informal wording, spelling and punctuation.

FIELDS
- Explanation: revise from the expert's answers.
- Examples: this field means "clearly IS this label". Keep and refine what the entry already has. You may add a contested turn ONLY where the expert's answers place it squarely inside the category --- not merely inside the argument about it. These turns were selected because readers could not agree on them, so most will not clear that bar; adding none is a perfectly good outcome. Never invent an example to fill the field.
- Near-Hit (close but not quite), Near-Miss (yes, but almost not) and Non-Examples: these are about boundaries, and a turn readers split on IS a boundary case. Where these fields are EMPTY, treat that as a gap to fill, not as text to preserve --- they are the largest hole in this codebook. Fill them with contested turns quoted verbatim, choosing ones the expert's answers place on that side of the line. Each entry must carry a short tag in parentheses naming the category it actually belongs to, e.g. "(OTHER_LABEL)", or what makes it marginal. Separate entries with ' | '. If the answers do not settle which side a turn falls on, leave it out.

Also:
- Keep the original voice and level of detail. This is a working codebook, not an essay.
- Keep every field that already exists. Do not rename the label or change its category.
- Never write an example you cannot point to in the contested turns, and never write a rule you cannot point to in an answer.

Reply with a single JSON object and nothing else:
{"revised": {"<field>": "<full text of the field after revision>", ...},
 "changed_fields": ["<field>", ...],
 "grounded_in": {"<field>": ["Q1", "Q3"], ...},
 "rationale": "<2-4 sentences: what you changed and which answers drove it>"}
"revised" must contain exactly the field names you were given. "grounded_in" must name, for every field you changed, the question numbers whose answers justify the change.

=============================== USER PROMPT ===============================
You are revising the codebook entry for tutoring-move label **{label}**.

No expert ground truth is available. 

=== ORIGINAL CODEBOOK ENTRY ===
LABEL: {label}
CATEGORY: {category}

## Current definition
{target_label_definition}

## Ambiguities diagnosed in this entry 
{ambiguities_patterns}

## Expert answers
{expert_questions_answers}

## Fields to return
["Explanation", "Examples", "Near-Miss (yes, but almost not)", "Near-Hit (close but not quite)", "Non-Examples"]

Revise the entry and reply with the JSON object described in your instructions.
\end{promptlisting}
\captionof{figure}{System and user prompt templates for the \textbf{Ambiguity-pattern + QA} revision condition. The diagnosis, the competing categories and the contested turns are supplied as context, with the expert's answers marked as the only source of rules.}
\Description{System and user prompt templates for revising codebook entries using the Ambiguity-pattern and expert clarification answers
}
\label{fig:qa-revision-ambiguity-qa}

\begin{promptlisting}{QA codebook revision: Ambiguity Pattern + within- and cross-label QA (1/2)}
============================== SYSTEM PROMPT ==============================

You are a methodologist maintaining a qualitative-coding codebook. You are revising ONE label whose definition was found to be ambiguous. You are given the label's current entry, a diagnosis of the ambiguities found in it, the categories readers chose instead, real turns readers disagreed about, and expert testimony in two forms. Rewrite the entry so the ambiguities are resolved.

These sources do NOT carry equal authority.

- EXPERT DIRECT ANSWERS are answers to questions asked about THIS label. They are direct testimony and the strongest source of rules.
- EXPERT CROSS-LABEL ANSWERS are answers to questions asked about a DIFFERENT label, in which the expert nonetheless ruled on this one --- usually while drawing the line between the two. Each is shown with the label it was asked about. Find what the answer says about THIS label and treat only that as binding. The rest of the answer is testimony about the other label: it tells you where the boundary runs, and it must NOT be imported into this entry as though it described this category.
- THE DIAGNOSIS says what each question was asking and why readers split. Use it to interpret a terse answer. It is analysis, not evidence, and never a source of rules.
- THE CONTESTED TURNS are real utterances from the corpus, and they are the SAME turns the questions were written from. An answer that says "a bare request can count" is a ruling about the kind of turn shown here, so read the answers against these utterances rather than in the abstract. They are also the source of example TEXT: quote them verbatim, keeping their informal wording, spelling and punctuation.

Some labels have NO expert direct answers and are revised entirely from cross-label ones. There the evidence is real but narrow: an expert drawing a line said something about this side of it. Write what that supports and no more. A single sharpened sentence grounded in one ruling is the right outcome; a full rewrite is not.

FIELDS
- Explanation: revise from the expert's answers.
- Examples: this field means "clearly IS this label". Keep and refine what the entry already has. You may add a contested turn ONLY where the expert's answers place it squarely inside the category --- not merely inside the argument about it. These turns were selected because readers could not agree on them, so most will not clear that bar; adding none is a perfectly good outcome. Never invent an example to fill the field.
- Near-Hit (close but not quite), Near-Miss (yes, but almost not) and Non-Examples: these are about boundaries, and a turn readers split on IS a boundary case. Where these fields are EMPTY, treat that as a gap to fill, not as text to preserve --- they are the largest hole in this codebook. Fill them with contested turns quoted verbatim, choosing ones the expert's answers place on that side of the line. Each entry must carry a short tag in parentheses naming the category it actually belongs to, e.g. "(OTHER_LABEL)", or what makes it marginal. Separate entries with ' | '. If the answers do not settle which side a turn falls on, leave it out.

Expert cross-label answers are especially good evidence for these three fields: an expert saying "X is a simple statement of accuracy, Y is about how the student got there" is telling you what belongs on each side of that line.

Also:
- Keep the original voice and level of detail. This is a working codebook, not an essay.
- Keep every field that already exists. Do not rename the label or change its category.
- Never write an example you cannot point to in the contested turns, and never write a rule you cannot point to in an answer.

Reply with a single JSON object and nothing else:
{"revised": {"<field>": "<full text of the field after revision>", ...},
 "changed_fields": ["<field>", ...],
 "grounded_in": {"<field>": ["Q1", "C3"], ...},
 "rationale": "<2-4 sentences: what you changed and which answers drove it>"}
"revised" must contain exactly the field names you were given. "grounded_in" must name, for every field you changed, the answer IDs that justify it --- Q_n for a direct answer, C_n for a cross-label one.
\end{promptlisting}
\captionof{figure}{System prompt template for Question QA revision in the \textbf{Ambiguity-pattern + within- and cross-label QA} condition, which supplies expert testimony in two forms and ranks them by authority.}
\Description{System prompt for revising codebook entries using ambiguity patterns and expert direct and cross-label answers, ranked by authority.}
\label{fig:qa-revision-cross-qa-system}

\begin{promptlisting}{QA codebook revision: Ambiguity Pattern + within- and cross-label QA (2/2)}
=============================== USER PROMPT ===============================
You are revising the codebook entry for tutoring-move label **{label}**.

No expert ground truth is available. 

=== ORIGINAL CODEBOOK ENTRY ===
LABEL: {label}
CATEGORY: {category}

## Current definition
{target_label_definition}

## Ambiguities diagnosed in this entry
{ambiguities_patterns}

## Expert direct answers about {label}
{expert_questions_direct_answers}

## Expert cross-label answers that ruled on {label}
{expert_cross_questions_label_answers}

## Fields to return
["Explanation", "Examples", "Near-Miss (yes, but almost not)", "Near-Hit (close but not quite)", "Non-Examples"]

Revise the entry and reply with the JSON object described in your instructions.
\end{promptlisting}
\captionof{figure}{User prompt template for the \textbf{Ambiguity-pattern + within- and cross-label QA} revision condition.}
\Description{User prompt for revising codebook entries using ambiguity patterns and within- and cross-label expert answers, with direct answers given the highest authority.}
\label{fig:qa-revision-cross-qa-user}

\subsection{Rationale Labeling}

\paragraph{Collecting Expert Labels and Rationales.}
We selected 120 utterances across the 12 target labels, ten per label, using the procedure described above (Section~\ref{app:sec-utt-selection}).
Each expert's 60 assigned utterances were pooled and shuffled before presentation. Experts saw each utterance in its surrounding conversation, but neither the target label used for selection nor model predictions.
They could assign any codebook label and were required to provide a written rationale, yielding 120 annotations with rationales in total.

\paragraph{Revising the Codebook from Expert Annotations.}
We grouped annotated utterances by their expert-assigned labels. For each represented label, the revision prompt supplied its Initial Codebook entry, the utterances assigned to it, their surrounding conversation, and expert rationales (Figure~\ref{fig:rl-revision}).
The prompt specified that new criteria, boundaries, and exclusions must be grounded in these rationales. 

\begin{promptlisting}{RL codebook revision}
============================== SYSTEM PROMPT ==============================

You are an expert in educational research, learning science, and tutoring codebook design.

Your job is to produce a *conceptual* revision of a tutoring-move codebook entry using expert annotation evidence.

The evidence is a set of ASSIGNED UTTERANCES: turns an expert annotator placed under this label, each with the surrounding conversation and the reason they gave for choosing it.

The reasons are the only source of rules. A criterion, boundary or exclusion must come from something an expert wrote; the bare fact that an utterance was assigned here is not itself a reason. Where the reasons do not settle something, leave it unsettled. Where two experts imply different rules, say so in the entry rather than picking one silently.

These utterances were selected because independent readers disagreed about them, and were shown shuffled with the original label hidden. They are HARD CASES judged blind, not a representative sample: sharpen the entry against them rather than widening it to accommodate every one.

FIELDS
- Explanation: revise from the experts' reasons.
- Examples: this field means "clearly IS this label". Keep and refine what the entry already has. You may add a contested turn ONLY where a reason places it squarely inside the category --- not merely inside the argument about it. These turns were selected because readers could not agree on them, so most will not clear that bar; adding none is a perfectly good outcome. Never invent an example to fill the field.
- Near-Hit (close but not quite), Near-Miss (yes, but almost not) and Non-Examples: these are about boundaries, and a turn readers split on IS a boundary case. Where these fields are EMPTY, treat that as a gap to fill, not as text to preserve --- they are the largest hole in this codebook. Fill them with contested turns quoted verbatim, choosing ones the reasons place on that side of the line --- a reason often rules something out while ruling this label in, and that is what these fields are for. Each entry must carry a short tag in parentheses naming the category it actually belongs to, e.g. "(OTHER_LABEL)", or what makes it marginal. Separate entries with ' | '. If the reasons do not settle which side a turn falls on, leave it out.
- Keep every field that already exists, keep the original voice, and do not rename the label or change its category.

Reply with a single JSON object and nothing else:
{"revised": {"<field>": "<full text of the field after revision>", ...},
 "changed_fields": ["<field>", ...],
 "grounded_in": {"<field>": ["E1", "M4"], ...},
 "rationale": "<2-4 sentences: what you changed and which reasons drove it>"}
"revised" must contain exactly the field names you were given. "grounded_in" must name, for every field you changed, the evidence ids whose reasons justify it.

=============================== USER PROMPT ===============================
You are revising the codebook entry for tutoring-move label **{label}**.

No expert ground truth is available. 

=== ORIGINAL CODEBOOK ENTRY ===
LABEL: {label}
CATEGORY: {category}

## Current definition
{target_label_definition}

## Utterances the experts ASSIGNED to {label}  
{utterances_the_experts_assigned_to}

## Fields to return
["Explanation", "Examples", "Near-Miss (yes, but almost not)", "Near-Hit (close but not quite)", "Non-Examples"]

Revise the entry and reply with the JSON object described in your instructions.
\end{promptlisting}

\captionof{figure}{System and user prompt templates for \textbf{RL-revision}. The model is given the utterances an expert assigned to the label, each with its surrounding conversation and the reason the expert gave, and those rationales are the only permitted source of criteria.}
\Description{System and user prompts for RL codebook revision using expert-labeled utterances and conversation context, with expert rationales as the sole source of revision rules.
}
\label{fig:rl-revision}

\subsection{Codebook Evaluation and Prediction Aggregation}
We evaluated each resulting codebook on the held-out set of 4,032 expert-consensus tutor utterances from 110 sessions. 
The same six models independently annotated this set using the shared utterance annotation prompt (Figure~\ref{fig:utterance-annotation-prompt}), with only the inserted codebook varying across conditions. 
Each model completed one run per condition at temperature~0, without persona conditioning. 

For each codebook, we combined the six models' predictions using eight Crowd-Kit aggregation methods~\cite{CrowdKit}: Majority Vote, Dawid--Skene, One-Coin Dawid--Skene, GLAD, MMSR, Wawa, Zero-Based Skill, and MACE. We calculated accuracy and weighted F1 against the expert-consensus labels separately for each aggregation method. 
Reported main-study scores are arithmetic means across these eight methods. When an expert-consensus annotation contained multiple labels, evaluation used the first recorded label.



\section{Additional Analysis of Expert Effort}\label{app:additional-analysis-effort}
\begin{figure*}
    \centering
    \includegraphics[width=0.9\linewidth]{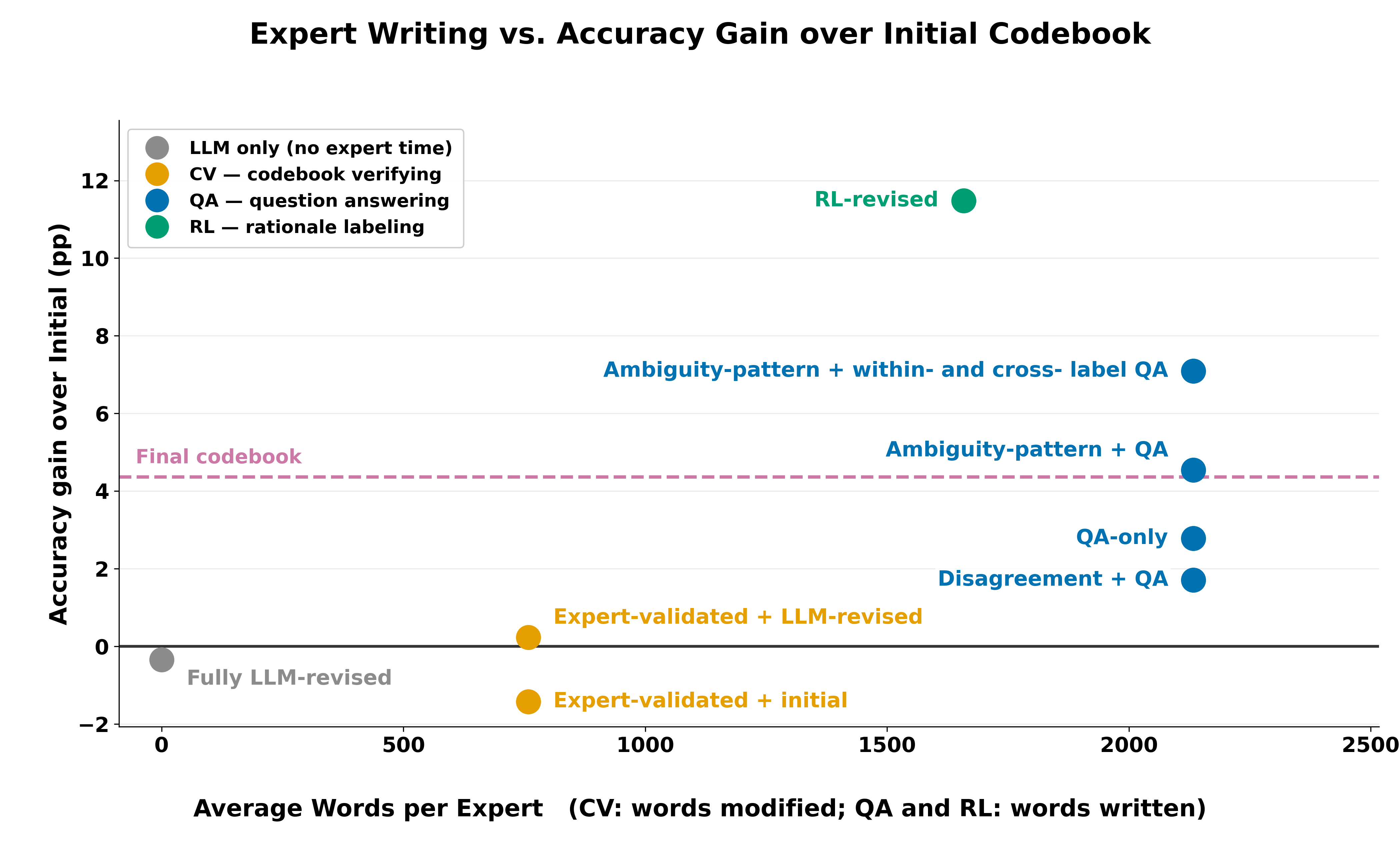}
    \caption{Expert written input and accuracy gains relative to the initial codebook. The x-axis shows average word counts across the two experts: modified words in codebook edits for CV, answer words for QA, and rationale words for RL. The y-axis shows accuracy gains in percentage points, averaged across eight aggregation methods. Each point represents a codebook condition, colored by workflow. The solid line marks zero gain; the dashed line indicates the final codebook's gain. RL achieved the largest accuracy improvement with fewer written words than QA.}
    \Description{Scatter plot of average words per expert versus accuracy gain over the initial codebook. RL achieves the largest gain, approximately 11.5 percentage points, with fewer words than QA (1,658 versus 2,133). CV involves fewer modified words but yields little improvement or decreased accuracy. A dashed line marks the final codebook's gain.}
    \label{fig:word-vs-acc}
\end{figure*}

Table~\ref{tab:expert-effort} reports each expert's time and written contributions across the three workflows. 
Using average word count per expert as a complementary proxy for effort, Figure~\ref{fig:word-vs-acc} shows a similar overall pattern to the time-based analysis in Figure~\ref{fig:effort-accuracy}: 
RL achieves the largest accuracy gain with fewer words of expert input than QA, whereas CV involves the fewest words but yields little or no improvement over the Initial Codebook. Word counts reflect words modified in CV, answer words in QA, and rationale words in RL.

\begin{table*}[t]
\centering

\begin{tabular}{llrr}
\toprule
\textbf{Workflow (written input)}&
\textbf{Expert} &
\textbf{Time (min)} &
\textbf{Total Words} \\
\midrule
\multirow{3}{*}{CV (codebook edits)}& A & 25.6
& 223\\
   & B & 80.5& 1,293\\
 & Mean& 53.1& 758\\
\midrule
\multirow{3}{*}{QA (answers)}& A & 222.5
& 1,791\\
   & B & 136.0
& 2,475\\
 & Mean& 179.2& 2,133\\
\midrule
\multirow{3}{*}{RL (annotation rationales)} & A & 146.9
&  1,340\\
   & B & 191.1
&  1,976\\
 & Mean& 169.0& 1,658\\
 \bottomrule
\end{tabular}
\caption{Expert effort across the CV, QA, and RL workflows. Word counts reflect modified words for CV, answer words for QA, and rationale words for RL. For CV, word counts include only cases in which experts made edits. 
Figures~\ref{fig:effort-accuracy} and~\ref{fig:word-vs-acc} relate average expert time and word counts, respectively, to accuracy gains over the initial codebook.}
\Description{Table reporting time and word counts for two experts across CV, QA, and RL, with workflow means. CV has the lowest average time and word count (53.1 minutes; 758 words), followed by RL (169.0 minutes; 1,658 words) and QA (179.2 minutes; 2,133 words).
}
\label{tab:expert-effort}
\end{table*}

\section{Post-Task Survey}
\label{app:survey}

\subsection{Survey Instrument}

\textbf{Post-Task Survey: Reflections on LLM-Supported Codebook Revision Workflows}

\medskip
Thank you for your time and effort on the LLM codebook revision task! We would like to hear your reflections on the process.
 
There are no right or wrong answers. Please respond based on your own experience and honest opinion.
 
\medskip
\noindent\textit{The Three Workflows}
 
\medskip
\noindent During this task, you completed three different workflows designed to support LLM codebook revision:
 
\begin{description}
    \item[S1 --- Codebook Verification] Reviewing and editing LLM-proposed revisions to the definitions or descriptions of selected labels.
    \item[S2 --- Question Answering] Answering LLM-generated clarification questions about label definitions, distinctions, and boundaries.
    \item[S3 --- Rationale Labeling] Labeling specific utterances and explaining the reasoning behind your labeling decisions.
\end{description}

\noindent For the questions below, please rate each workflow separately.
 
\bigskip
\noindent\textbf{Your Name} *

\bigskip
\noindent\textbf{1. This workflow required substantial mental effort from me. *}

\smallskip
\noindent\textit{Strongly Disagree - 1 \quad 2 \quad 3 \quad 4 \quad 5 \quad 6 \quad 7 - Strongly Agree}
 
\smallskip
\noindent S1 --- Codebook Verification
 
\noindent S2 --- Question Answering
 
\noindent S3 --- Rationale Labeling
 
\bigskip
\noindent\textbf{2. This workflow prompted me to think carefully about how labeling decisions should be made. *}

\smallskip
\noindent\textit{Strongly Disagree - 1 \quad 2 \quad 3 \quad 4 \quad 5 \quad 6 \quad 7 - Strongly Agree}
 
\smallskip
\noindent S1 --- Codebook Verification
 
\noindent S2 --- Question Answering
 
\noindent S3 --- Rationale Labeling
 
\bigskip
\noindent\textbf{3. This workflow required me to repeat work without contributing useful new information. *}

\smallskip
\noindent\textit{Strongly Disagree - 1 \quad 2 \quad 3 \quad 4 \quad 5 \quad 6 \quad 7 - Strongly Agree}
 
\smallskip
\noindent S1 --- Codebook Verification
 
\noindent S2 --- Question Answering
 
\noindent S3 --- Rationale Labeling
 
\bigskip
\noindent\textbf{4. This workflow allowed me to express the prior knowledge I already had about distinctions between labels. *}
 
\smallskip
\noindent\textit{Strongly Disagree - 1 \quad 2 \quad 3 \quad 4 \quad 5 \quad 6 \quad 7 - Strongly Agree}
 
\smallskip
\noindent S1 --- Codebook Verification
 
\noindent S2 --- Question Answering
 
\noindent S3 --- Rationale Labeling
 
\bigskip
\noindent\textbf{5. This workflow helped me notice meaningful distinctions between labels that I had not previously considered. *}

\smallskip
\noindent\textit{Strongly Disagree - 1 \quad 2 \quad 3 \quad 4 \quad 5 \quad 6 \quad 7 - Strongly Agree}
 
\smallskip
\noindent S1 --- Codebook Verification
 
\noindent S2 --- Question Answering
 
\noindent S3 --- Rationale Labeling
 
\bigskip
\noindent\textbf{6. This workflow failed to capture important prior knowledge I already had about distinctions between labels. *}

\smallskip
\noindent\textit{Strongly Disagree - 1 \quad 2 \quad 3 \quad 4 \quad 5 \quad 6 \quad 7 - Strongly Agree}
 
\smallskip
\noindent S1 --- Codebook Verification
 
\noindent S2 --- Question Answering
 
\noindent S3 --- Rationale Labeling
 
\bigskip
\noindent\textbf{7. I expect that a codebook revised using this workflow would help an LLM annotate new utterances accurately. *}

\smallskip
\noindent\textit{Strongly Disagree - 1 \quad 2 \quad 3 \quad 4 \quad 5 \quad 6 \quad 7 - Strongly Agree}
 
\smallskip
\noindent S1 --- Codebook Verification
 
\noindent S2 --- Question Answering
 
\noindent S3 --- Rationale Labeling
 
\bigskip
\noindent\textbf{8. Overall, I liked using this workflow. *}

\smallskip
\noindent\textit{Strongly Disagree - 1 \quad 2 \quad 3 \quad 4 \quad 5 \quad 6 \quad 7 - Strongly Agree}
 
\smallskip
\noindent S1 --- Codebook Verification
 
\noindent S2 --- Question Answering
 
\noindent S3 --- Rationale Labeling
 
\bigskip
\noindent\textbf{9. Is there anything else you would like to share with us about the task or your experience with the three workflows?}
 
\subsection{Survey Results}
 
Table~\ref{tab:survey-results} reports ratings from two domain experts (Expert A, Expert B) for each item across the three workflows, along with the average of the two ratings. All ratings use the 7-point scale described above; for item 3 (repetition) and item 6 (failure to capture prior knowledge), lower scores are more favorable, while for all other items, higher scores are more favorable.
 
\begin{table*}
\centering
\small
\begin{tabular}{@{}p{6.2cm}lccc@{}}
\toprule
\textbf{Item} & \textbf{Workflow} & \textbf{Expert A} & \textbf{Expert B} & \textbf{Avg.} \\
\midrule
 
\multirow{3}{6.2cm}{Required substantial mental effort}
 & S1 -- Codebook Verification & 5 & 5 & 5.0 \\
 & S2 -- Question Answering    & 7 & 7 & 7.0 \\
 & S3 -- Rationale Labeling    & 4 & 3 & 3.5 \\
\midrule
 
\multirow{3}{6.2cm}{Prompted careful thinking about labeling decisions}
 & S1 -- Codebook Verification & 5 & 7 & 6.0 \\
 & S2 -- Question Answering    & 7 & 7 & 7.0 \\
 & S3 -- Rationale Labeling    & 7 & 7 & 7.0 \\
\midrule
 
\multirow{3}{6.2cm}{Required repeating work without new information}
 & S1 -- Codebook Verification & 3 & 3 & 3.0 \\
 & S2 -- Question Answering    & 1 & 1 & 1.0 \\
 & S3 -- Rationale Labeling    & 2 & 2 & 2.0 \\
\midrule
 
\multirow{3}{6.2cm}{Allowed expression of prior knowledge about label distinctions}
 & S1 -- Codebook Verification & 3 & 5 & 4.0 \\
 & S2 -- Question Answering    & 4 & 7 & 5.5 \\
 & S3 -- Rationale Labeling    & 7 & 7 & 7.0 \\
\midrule
 
\multirow{3}{6.2cm}{Helped notice previously unconsidered distinctions}
 & S1 -- Codebook Verification & 3 & 7 & 5.0 \\
 & S2 -- Question Answering    & 7 & 7 & 7.0 \\
 & S3 -- Rationale Labeling    & 4 & 7 & 5.5 \\
\midrule
 
\multirow{3}{6.2cm}{Failed to capture important prior knowledge}
 & S1 -- Codebook Verification & 1 & 4 & 2.5 \\
 & S2 -- Question Answering    & 1 & 4 & 2.5 \\
 & S3 -- Rationale Labeling    & 1 & 4 & 2.5 \\
\midrule
 
\multirow{3}{6.2cm}{Expected to help an LLM annotate new utterances accurately}
 & S1 -- Codebook Verification & 5 & 5 & 5.0 \\
 & S2 -- Question Answering    & 7 & 7 & 7.0 \\
 & S3 -- Rationale Labeling    & 7 & 6 & 6.5 \\
\midrule
 
\multirow{3}{6.2cm}{Overall liked using this workflow}
 & S1 -- Codebook Verification & 7 & 6 & 6.5 \\
 & S2 -- Question Answering    & 7 & 7 & 7.0 \\
 & S3 -- Rationale Labeling    & 7 & 4 & 5.5 \\
 
\bottomrule
\end{tabular}
\caption{Post-task survey ratings by workflow (1 = Strongly Disagree, 7 = Strongly Agree).}\label{tab:survey-results}
\end{table*}
 
Two experts' responses to the last open-ended question are as follows.
 
\begin{quote}
\itshape
Expert A: ``I am amazed at how deeply the questions made me think. That was an incredible experience considering I helped write/refine the original code book. I appreciated the Rationale Labeling. It was meaningful to me because I needed to be able to fully justify my thinking. It made me think carefully about code distinctions. The verification process felt very similar to what our team had done since September 2025. The platform was very user friendly.''
\end{quote}

\begin{quote}
\itshape
Expert B: ``I really enjoyed *Labeling* from the workflows... by identifying and comparing codes, and answering questions about them, I really feel like my understanding of the nuances of each code grew. This happened over time, so the quality of my responses changed as I worked.''
\end{quote}

\end{document}